\documentclass[runningheads]{llncs}

\usepackage{eccv}

\usepackage{eccvabbrv}

\usepackage{graphicx}
\usepackage{booktabs}
\usepackage{threeparttable}
\usepackage{makecell}
\usepackage{array,multirow}
\usepackage{setspace}
\usepackage[export]{adjustbox}
\usepackage{pifont}
\usepackage[accsupp]{axessibility}  
\newcommand{\repeatthanks}{\textsuperscript{\thefootnote}}

\usepackage{hyperref}

\hypersetup{
 colorlinks=true,
 linkcolor=red,
 citecolor=cyan,
 filecolor=magenta, 
 urlcolor=cyan,
 }
\usepackage{orcidlink}

\begin{document}

\title{One-DM: One-Shot Diffusion Mimicker for Handwritten Text Generation}

\titlerunning{One-DM: One-Shot Diffusion Mimicker for Handwritten Text Generation}


\author{Gang Dai$^{1}$\thanks{Authors contributed equally; $^{\dagger}$ Corresponding author}\orcidlink{0000-0001-8864-908X} \and
Yifan Zhang$^{2,3}$\repeatthanks\orcidlink{0000-0002-2125-1074} \and \\
Quhui Ke\orcidlink{0000-0002-1393-3059}$^{1}$\and 
Qiangya Guo$^{1}$\orcidlink{0009-0001-5347-6198}\and
Shuangping Huang$^{1,4\dagger}$\orcidlink{0000-0002-5544-4544}}
\institute{$^{1}$South China University of Technology\\ 
$^{2}$National University of Singapore, $^{3}$Skywork AI, $^{4}$Pazhou Laboratory \\ \email{\{eedaigang@mail., eehsp@\}scut.edu.cn, yifan.zhang@u.nus.edu}}

\authorrunning{G. Dai et al.}


\maketitle

\begin{abstract}

Existing handwritten text generation methods often require more than ten handwriting samples as style references. However, in practical applications, users tend to prefer a handwriting generation model that operates with just a single reference sample for its convenience and efficiency. This approach, known as ``one-shot generation'', significantly simplifies the process but poses a significant challenge due to the difficulty of accurately capturing a writer's style from a single sample, especially when extracting fine details from the characters' edges amidst sparse foreground and undesired background noise. To address this problem, we propose a One-shot Diffusion Mimicker (One-DM) to generate handwritten text that can mimic any calligraphic style with only one reference sample. Inspired by the fact that high-frequency information of the individual sample often contains distinct style patterns (\eg, character slant and letter joining), we develop a novel style-enhanced module to improve the style extraction by incorporating high-frequency components from a single sample. We then fuse the style features with the text content as a merged condition for guiding the diffusion model to produce high-quality handwritten text images. Extensive experiments demonstrate that our method can successfully generate handwriting scripts with just one sample reference in multiple languages, even outperforming previous methods using over ten samples. Our source code is available at \url{https://github.com/dailenson/One-DM}.
\end{abstract}

\section{Introduction}
\label{sec:intro}
In the digital age, handwriting text generation blends the personalization of traditional handwriting with the efficiency of automated processes, offering a digital format to preserve the authenticity of individual handwriting. This task aims to automatically generate the desired handwritten text images that not only correspond to specific text content, but also emulate the calligraphic style of a given exemplar writer (e.g., character slant, cursive join, stroke thickness, and ink color). This provides great convenience for people with hand impairments and also contributes to accelerating the process of handwritten font design.

Previous works~\cite{fogel2020scrabblegan,kang2020ganwriting,bhunia2021handwriting,pippi2023handwritten,DavisMPTWJ20,gan2021higan,gan2022higan+} introduce generative adversarial networks (GANs)~\cite{goodfellow2014,cao2019multi,liu2021deep} for handwritten text generation. For instance, ScrabbleGAN~\cite{fogel2020scrabblegan} leverages random noises as style inputs and conditions content inputs on character-level labels, enabling the synthesis of handwritten words with randomly sampled styles.  In a recent study~\cite{dhariwal2021diffusion}, diffusion models~\cite{rombach2022high,zhang2023hipa,liu2023geom} such as denoising diffusion probabilistic models (DDPM)~\cite{ho2020denoising} have demonstrated even higher quality of image generation compared to GANs. This motivates several attempts, such as WordStylist~\cite{icdar_NikolaidouRCSSSML23}, GC-DDPM~\cite{Ding2023ImprovingHO}, and CTIG-DM~\cite{zhu2023conditional}, to condition the denoising process on the fixed writer ID to generate handwritten text images with controllable styles. However, the major limitation of these methods~\cite{fogel2020scrabblegan,zhu2023conditional,icdar_NikolaidouRCSSSML23,Ding2023ImprovingHO} is that they are unable to mimic the unseen writers' handwriting styles.

\begin{figure}[t]
  \centering 
  \includegraphics[width=0.9\linewidth]{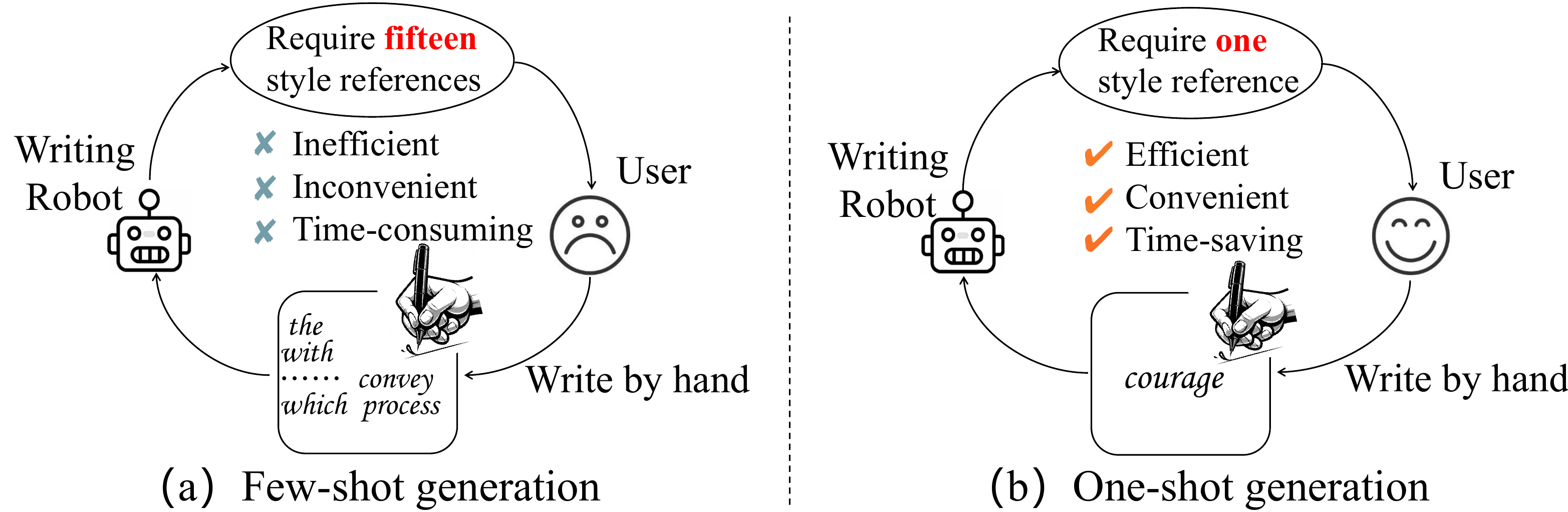}
 
   \caption{User experience comparisons between one-shot and few-shot handwritten text generation methods. It reveals that one-shot setting leads to a better user experience.}
\vspace{-0.2in}
   \label{fig:product}
\end{figure}

To imitate any given handwriting style, some previous methods~\cite{kang2020ganwriting,bhunia2021handwriting,pippi2023handwritten}, known as few-shot generation, require users to provide a few samples (typically 15) as style references. They employ a style encoder to extract writing styles from the given samples, thus offering flexible control over the style of the generated samples. 
However, the traditional few-shot generation pipeline is inconvenient, inefficient, and time-consuming, as shown in~\cref{fig:product}. Users prefer methods that only require a single sample as a style reference, known as one-shot generation methods, because they are more convenient, efficient, and easy-to-use. Our goal is to investigate the more challenging one-shot generation task that holds significant practical value, striving to achieve high-quality handwritten text image generation with desired styles and contents.

\begin{figure}[t]
  \centering
  \includegraphics[width=0.73\linewidth]{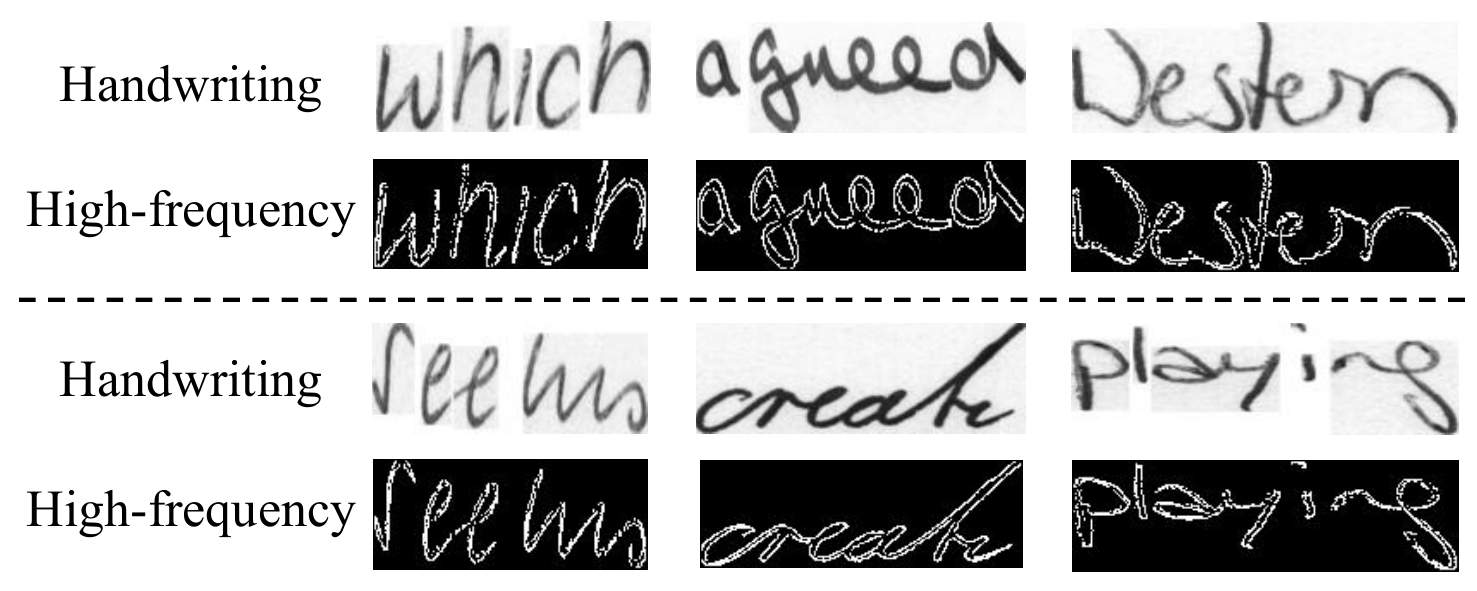}

   \caption{
Handwritten text samples and corresponding high-frequency components. We find that high-frequency components have more pronounced character contours, clearly showcasing the style patterns, such as character slant and cursive connections.}
\vspace{-0.2in}
   \label{fig:samples_highf_requency}
\end{figure}

The primary challenge of one-shot generation task lies in accurately extracting a user's handwriting style from just one style reference image. As illustrated in~\cref{fig:samples_highf_requency}, characters, being abstract symbols, typically occupy only a minor portion of the reference image. Besides, the reference image is often cluttered with noisy background information, which poses a significant obstacle in extracting the individual handwriting style. Previous one-shot generation methods~\cite{DavisMPTWJ20,gan2021higan,gan2022higan+} simply follow the architectural design of few-shot approaches, using a vanilla CNN encoder to directly extract the handwriting style from a single sample. The extracted style is then combined with text contents and input into a CNN decoder to generate the desired handwritten images. These methods exhibit a limited performance in emulating handwriting styles, due to their poor ability of style extraction. 


To address the above challenges, our key idea revolves around utilizing the high-frequency information of the sample to enhance the extraction of the handwriting style. As depicted in~\cref{fig:samples_highf_requency}, the high-frequency information encompasses the overall contours of the handwritten text, allowing for a clear observation of key style patterns such as the text slant, letter spacing, and cursive connections. Hence, the incorporation of high-frequency information facilitates a more effective handwriting style extraction.

In light of the above insight, we propose a One-shot Diffusion Mimicker (One-DM) for handwritten text generation, which is simultaneously guided by desired style and arbitrary content. More specifically, we first develop a style-enhanced module to process both a style reference image and its high-frequency components in parallel. Considering that the reference image often contains background noise, we design a gate mechanism to suppress the inflow of background noise. Regarding the high-frequency components, such as character slant and ligature that show clearer style patterns, we employ a contrastive learning framework~\cite{khosla2020supervised,zhang2021unleashing,zhang2022self,lin2024contrastive} to further obtain discriminative style features, guiding the handwritten text synthesis with both realistic and diverse styles. Subsequently, the style features extracted from both branches are adaptively fused with the specific content prototype in a style-content fusion module. Finally, the seamlessly integrated style and content features act as conditional inputs, guiding the denoising process for the progressive synthesis of stylized handwritten text images. It is worth noting that our One-DM effectively mimics a user's writing style with only one reference sample through our style-enhanced module, surpassing few-shot methods in producing higher-quality stylized handwritings.

To sum up, our contributions are as follows:
\begin{itemize}
\item We propose a novel diffusion model for stylized handwritten text generation, which only requires a single reference sample as style input, and imitates its writing style to generate handwritten text with arbitrary content.
\item We introduce the high-frequency components of the reference sample to enhance the extraction of handwriting style. The proposed style-enhanced module can effectively capture the writing style patterns and suppress the interference of background noise.
\item Extensive experiments on handwriting datasets in English, Chinese, and Japanese demonstrate that our approach with a single style reference even outperforms previous methods with 15x-more references.
\end{itemize}

\section{Related Work}
\textbf{Handwriting generation.} 
Handwritten text is typically stored in two formats: online trajectory form or offline image form. Online handwriting generation methods~\cite{aksan2018deepwriting,kotani2020generating,tolosana2021deepwritesyn,zhao2020deep,tang2021write,dai2023disentangling} often employ Recurrent Neural Networks (RNNs)~\cite{aksan2018deepwriting,kotani2020generating,tolosana2021deepwritesyn,zhao2020deep,tang2021write,tang2019fontrnn,  chen2022complex}, transformer decoders\cite{dai2023disentangling}, or diffusion models~\cite{luhman2020diffusion,ren2023diff} to progressively generate writing trajectories. Unlike online methods, offline generation methods have the advantage of not requiring additional trajectory supervision information and can generate realistic handwritten text with stroke width and ink color, which online methods cannot produce.

Previous offline handwriting generation methods~\cite{alonso2019adversarial,fogel2020scrabblegan,kang2020ganwriting,bhunia2021handwriting,pippi2023handwritten,DavisMPTWJ20,luo2022slogan,huang2022agtgan} in deep learning predominantly rely on Generative Adversarial Networks (GANs). Early works~\cite{alonso2019adversarial,fogel2020scrabblegan} condition the generative process on the word embeddings~\cite{alonso2019adversarial} or concatenated letter-tokens~\cite{fogel2020scrabblegan} to synthesize handwritten word images. However, these methods struggle to flexibly control the writing style. Thus, few-shot approaches~\cite{kang2020ganwriting,bhunia2021handwriting,pippi2023handwritten} which rely on 15 style references are introduced. For instance, GANwriting~\cite{kang2020ganwriting} utilizes a CNN encoder~\cite{he2016deep,yang2023cross,yang2024hilo} to extract a user's handwriting style from a few samples, which are then combined with specific text content to generate handwritings in the desired style. In its follow-up work~\cite{kang2021content}, the synthesized samples are demonstrated to assist in training more robust handwritten text recognizers. Further, HWT~\cite{bhunia2021handwriting} employs a transformer encoder~\cite{vaswani2017attention,li2024face} to extract rich style patterns from reference samples, enhancing the performance in style mimicry. Recently, VATr~\cite{pippi2023handwritten} uses images of symbols as content representations, enabling the generation of out-of-charset characters.

Concurrently, one-shot generation methods~\cite{DavisMPTWJ20,gan2021higan,gan2022higan+} are proposed. Despite their ability to mimic handwriting styles with only a single sample, these techniques still lag behind few-shot methods in terms of stylized generation results. Additionally, previous handwriting generation methods~\cite{alonso2019adversarial,fogel2020scrabblegan,kang2020ganwriting,bhunia2021handwriting, DavisMPTWJ20,gan2021higan,gan2022higan+} often rely on complex content representations, such as recurrent embeddings~\cite{alonso2019adversarial} and letter-level tokens~\cite{fogel2020scrabblegan,kang2020ganwriting,bhunia2021handwriting, DavisMPTWJ20,gan2021higan,gan2022higan+}. SLOGAN~\cite{luo2022slogan} extracts textual contents from easily obtainable printed images. However, it faces challenges in generalizing to unseen writing styles due to its fixed writer ID. Similarly, some diffusion-based~\cite{zhu2023conditional, icdar_NikolaidouRCSSSML23,Ding2023ImprovingHO} methods condition the denoising process on fixed style ID and are unable to mimic styles that they have not previously encountered. In contrast, our One-DM effectively
obtains style information from one style sample and thus can generate handwritings with arbitrary styles. Due to space constraints, we discuss more related works in~\cref{related}, including diffusion methods for general image generation, contrastive-based method~\cite{xie2022toward}, and frequency-based methods\cite{lin2021drafting,pan2023visa}.

\begin{figure*}
\vspace{-0.2in}
    \centering \includegraphics[width=0.95\linewidth]{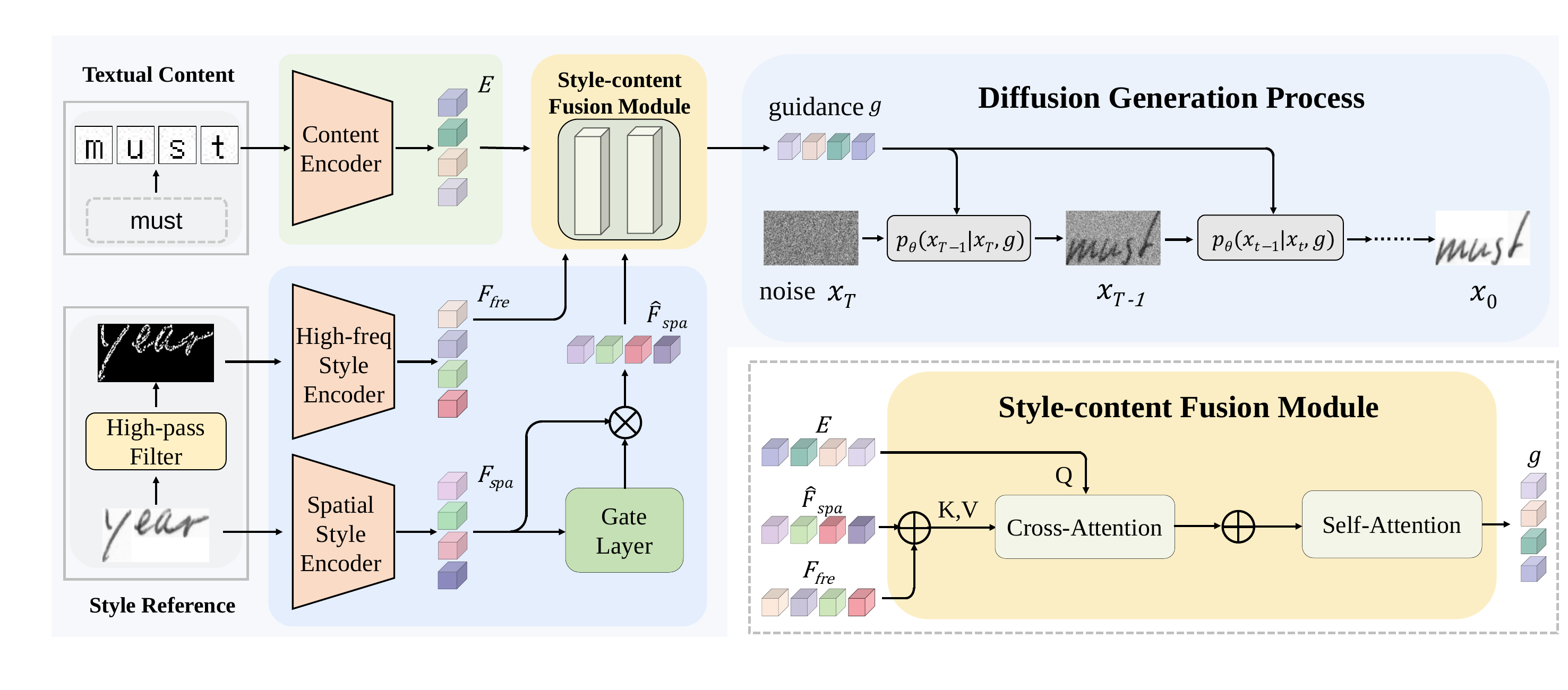}
    \vspace{-0.2in}
    \caption{ Overview of the proposed method. The style reference initially passes through a high-pass filter to extract its high-frequency components. Subsequently, the spatial and the high-frequency style encoders independently extract style features $F_{spa}$ and $F_{fre}$ from the style reference and its high-frequency information, respectively. $F_{spa}$, after being filtered through a gate mechanism, is fused with $F_{fre}$ and content features $E$ in the fusion module. The merged feature then serves as a condition input to guide the diffusion generation process.
    }
    \label{fig:method_overview}
    \vspace{-0.2in}
\end{figure*}

\section{Method}
\textbf{Problem statement.} We aim to synthesize handwritten text images that are controlled by both textual content and handwriting style. Given any word string $\mathcal{A}$ and a single style sample $I_s$ from a writer $w_s$, the generated handwritten word image $X_s$ should replicate the unique handwriting style of $w_s$ while maintaining the content of $\mathcal{A}$, where the textual content $\mathcal{A}\text{\small{=}}
\{a_i\}_{i=1}^L$ spans a length $L$ and $a_i$ is from a character set without any restrictions.

To address this task, we incorporate high-frequency information to enhance the extraction of writing styles. Existing methods~\cite{DavisMPTWJ20,kang2020ganwriting,gan2022higan+,gan2021higan,pippi2023handwritten,bhunia2021handwriting} typically use a vanilla CNN or transformer encoder to directly process style images, often resulting in undesired style extraction.
In contrast, we introduce a novel One-shot diffusion mimicker (One-DM) by innovating Laplacian high-frequency extraction and a gating mechanism. Our One-DM can effectively capture style features from a single reference while suppressing background noise.



\subsection{Overall Scheme}

Our high-level idea focuses on incorporating high-frequency information from style reference images to enhance the extraction of style patterns. A straightforward implementation involves using a vanilla transformer encoder to extract style features from both the style image and its corresponding high-frequency image. This naive solution encounters two main issues: (1) the lack of efficient supervision objective still makes it challenging to accurately learn the writer's style patterns from the high-frequency image, and (2) the style features captured from the original image still retain undesirable noise backgrounds, which can adversely affect subsequent image generation performance. 

To address the above-mentioned issues, we develop a more effective method, shown~\cref{fig:method_overview}. Our method consists of a style-enhanced module, a content encoder, a style-content fusion module, and a conditional diffusion module. Initially, we utilize the Laplacian kernel as a high-frequency filter to extract the high-frequency components from the style reference. Then, two parallel style encoders are employed to simultaneously extract the corresponding style features from both style reference and its high-frequency information. Since undesirable background noise is often present in style references, we design a gate mechanism to facilitate the transmission of informative style information while mitigating noise transmission. The style patterns in the high-frequency components are comparatively cleaner and more distinct, facilitating the observation of individual styles, such as character slant. For the observation, we propose a contrastive learning objective, termed LaplacianNCE $\mathcal{L}_{lapNCE}$, to enforce the more discriminative style learning from high-frequency components.

Regarding content guidance, we render the given string $\mathcal{A}$ into a unifont image, as done in VATr~\cite{pippi2023handwritten}. Briefly, Unifont's key advantage is that it covers all Unicode characters, allowing our method to transform any user inputs into corresponding images. We further feed render results into a content encoder that combines a ResNet18~\cite{he2016deep} with a Transformer encoder. This process first involves using a ResNet18 to handle each character image in parallel, followed by concatenating them to form word sequence features. The Transformer encoder then processes these features to extract an informative content feature $E\text{\small{=}}\{e_i\}_{i=1}^{L}\in \mathbb{R}^{L \times c}$ with a global context, where $c$ is the channel dimension. After obtaining style features and content guidance, we seamlessly fusion them using a style-content fusion module. The fused results then guide the denoising process of the conditional diffusion model to synthesize the desired handwritten text images. The denoising process is supervised by the reconstruction loss $\mathcal{L}_{rec}$.

To summarize, the overall training objective of our method
combines all two loss functions:
$$\mathcal{L} = \mathcal{L}_{lapNCE} + \mathcal{L}_{rec}.$$ 
\subsection{Style-enhanced Module}
We propose the style-enhanced module to enhance style extraction by incorporating high-frequency components $H_s \in \mathbb{R}^{h \times w \times c} $ from a reference image $I_s \in \mathbb{R}^{h \times w \times c} $ since clearer style patterns are presented in high-frequency components like character slanting and shape connections (cf.~\cref{fig:samples_highf_requency}). As shown in~\cref{fig:method_overview}, we use a Laplacian kernel as a high-frequency filter to extract $H_s$ from $I_s$. The Laplacian kernel excels in extracting high-frequency information without the need for Fast Fourier Transform (FFT) and parametric separation in the frequency domain.
Then, two style encoders, $\mathcal{E}_{spa}$ and $\mathcal{E}_{fre}$, each a combination of CNN and transformer, process the $I_s$ and $H_s$, respectively. This independent processing leads to the extraction of distinct style features:$F_{spa} \text{\small{=}} \{f_{spa}^{i}\}_{i=1}^{d} \in \mathbb{R}^{d \times c}$ from $\mathcal{E}_{spa}$  and $F_{fre} \text{\small{=}} \{f_{fre}^{i}\}_{i=1}^{d} \in \mathbb{R}^{d \times c}$ from $\mathcal{E}_{fre}$, where $d=h \times w$. While structurally identical, $\mathcal{E}_{spa}$ and $\mathcal{E}_{fre}$ do not share weights with each other. Then, the proposed $\mathcal{L}_{lapNCE}$ forces $\mathcal{E}_{fre}$ to focus on extracting discriminative style features from $H_s$.  A gate mechanism is designed to selectively filter out background noise from the reference style features, allowing only meaningful style patterns to pass.

\textbf{Laplacian Contrastive Learning.} The goal of the proposed $\mathcal{L}_{lapNCE}$ is to guide the high-frequency style encoder $\mathcal{E}_{fre}$ in learning more discriminative style features from the high-frequency information. Thus, we propose to bring closer extracted style features $F_{fre}$ belonging to the same writer, while distancing those from different writers. We formulate our $\mathcal{L}_{lapNCE}$ as follows:

\begin{equation}\label{WriterNCE}
\mathcal{L}_{lapNCE}\text{\small{=}\small{-}}\frac{1}{N}\sum_{i \in M}\frac{1}{	\left|P\left(i\right)\right|}\sum_{p \in P\left(i\right)}\log \frac{\exp{\left({z_i\cdot z_p}/\tau\right)}}{\sum_{a \in A\left(i\right)}\exp{\left({z_i \cdot z_a}/\tau\right)}}.
\end{equation}
In detail, $i \in M\text{\small{=}}\{1 ... N\} $ is the index of any element in the mini-batch with size $N$ and $A\left(i\right)\text{\small{=}}M \backslash \{i\}$ is other indices distinct from $i$. $z_i$ is an anchor sample belonging to writer $w_i$ and $P(i)\text{\small{=}}\{p \in A\left(i\right):w_p\text{\small{=}}w_i\}$ is its in-batch positive sample set and the other $A\left(i\right) \backslash P\left(i\right)$ is its negative set. Here, $z\text{\small{=}}Proj(F_{fre})$, where $Proj$ is a learnable multi-layer perceptron (MLP), $\tau$ is a scalar temperature parameter and the $\cdot$ symbol denotes the inner product.

\textbf{Gate mechanism.}
As illustrated in~\cref{fig:samples_highf_requency}, the stroke areas of characters in reference images are typically sparse, with background noise interfering with the extraction of character style features. To address this challenge, we propose a gate mechanism to selectively filter the information of the sample $I_s$, as shown in~\cref{fig:method_overview}. Specifically, the extracted sample style features $F_{spa}$ are fed into a gate layer, consisting of a learnable fully connected layer and followed by a sigmoid activation, to obtain the corresponding gate units $W\text{\small{=}}\{w_i\}_{i=1}^{d}\in \mathbb{R}^{d}$. Each unit $w_i$ determines the pass rate for the corresponding $f_{spa}^{i}$, allowing for a higher pass rate where $w_i$ is larger. This design effectively enables informative style features $\hat{F}_{spa}\text{\small{=}}\{\hat{f}_{spa}^{i}\}_{i=1}^{d}$ to be extracted while suppressing extraneous background noise, where $\hat{f}_{spa}^{i}=f_{spa}^{i}\cdot w_i$.

\subsection{Style-content Fusion Module}
Upon acquiring textual content feature $E$ and two style features $\hat{F}_{spa}$ and $F_{fre}$, we integrate all features within two multi-head attention mechanisms to guide the denoising generation process of the diffusion model, as shown in~\cref{fig:method_overview}. Specifically, the first cross-attention module uses the textual content $E$ as queries to identify the most relevant style information in the style reference, thereby inferring the style attributes corresponding to each character. For instance, if the textual content is `a', it prioritizes searching for style features of characters like `a', `b', `d', `g' in the style reference, due to these characters appearing similar looped structures, implying more comparable style attributes. This process (cross-attention in~\cref{fig:method_overview}) is represented as:
\begin{equation}
O=Atten_{1}(Q_1=E, K_1=V_1=\hat{F}_{spa}+F_{fre}).
\end{equation}
Subsequently, we obtain the initial fusion embedding between content and style guidance by simply summing $O$ and $E$. The merged intermediate vector is then employed as the query, key, and value in the self-attention mechanism to facilitate comprehensive interaction of information. Finally, the blended embedding $g$ serves as the condition of the diffusion process. The second multi-head attention (self-attention in~\cref{fig:method_overview}) is defined as:
\begin{equation}
g=Atten_{2}(Q_2=K_2=V_2=O+E).
\end{equation}

\subsection{Conditional Diffusion Model}
The goal of the conditional diffusion model $p_\theta$ is to generate realistic images of handwritten text, guided by acquired conditions $g$. Specifically, as shown in~\cref{fig:method_overview}, under the guidance of $g$, $p_\theta$ executes a denoising generative process with $T$ steps, starting from a sampled Gaussian noise $x_T$ and progressively denoises to obtain the desired handwritten text $x_0$:
\begin{equation}p_{\theta}(x_0|g)\text{\small{=}}\int p_{\theta}(x_{0:T}|g)d_{x_{1:T}},
\end{equation}
\begin{equation}p_{\theta}(x_{0:T}|g)\text{\small{=}}p(x_T)\prod \limits_{t=1}^Tp_{\theta}(x_{t-1}|x_t, g),
\end{equation}
\begin{equation}p_{\theta}(x_{t-1}|x_t,g)\text{\small{=}}\mathcal N(x_{t-1};\mu_{\theta}(x_t,g,t),\Sigma_{\theta}(x_t,g, t)).
\end{equation}

The denoising process aims to learn how to reverse a predefined forward process, as described in DDPM~\cite{ho2020denoising}. The forward process is modeled as a fixed Markov chain, where noise conforming to a normal distribution is incrementally added to $x_{t-1}$ to derive $x_t$. This can be mathematically expressed as:
\begin{equation}
q(x_{t}|x_{t-1}) = \mathcal{N}(x_{t}; \sqrt{1-\beta_t}x_{t-1}, \beta_t\mathbf{I}),
\end{equation}
where the noise is characterized by a variance schedule $\beta_t$. During training, a variational bound on the maximum likelihood objective is applied to guide the generation process to recover $x_0$ from the standard Gaussian noise $x_T$ conditioned on $g$. We give the training objective as follows:
\begin{equation}
\mathcal{L}_{rec}\text{\small{=}}\mathbb{E}_{t,q} \left\| \mu_t(x_t,x_0)-\mu_{\theta}(x_t,g,t)\right\|_2^2,
\end{equation}where $\mu_t(x_t,x_0)$ is the mean of the forward process posterior $q(x_{t}|x_{t-1})$, which has a closed-form solution~\cite{ho2020denoising}. We put more details in~\cref{diffusion}.

\section{Experiments}
\subsection{Experimental Settings}
\textbf{Dataset.}
To evaluate our One-DM in handwritten text generation, we use the widely-used handwriting dataset IAM~\cite{marti2002iam} and CVL~\cite{kleber2013cvl}. IAM consists of 62,857 English word images from 500 unique writers. Following~\cite{kang2020ganwriting,bhunia2021handwriting,pippi2023handwritten}, we use words from 339 writers for training and the remaining 161 writers for testing. CVL includes texts by 310 writers in English and German. We use the English portion, totaling 84,514 words, and follow CVL's standard split, with 283 writers for training and 27 for testing.
Throughout all experiments, we resize images to 64 pixels in height, preserving their aspect ratio.

\textbf{Evaluation metrics.}
We use the Fréchet Inception Distance (FID)~\cite{heusel2017gans} and Geometry Score (GS)~\cite{khrulkov2018geometry} to assess the generation quality, following the settings of~\cite{kang2020ganwriting,bhunia2021handwriting,pippi2023handwritten}. 
We also conduct user studies to quantify the subjective quality of the generated handwritten text images in~\cref{user_study}.

\textbf{Implementation details.} In our all experiments, we only use single style reference sample. 
We first train our model for $700$ epochs (batch size of 384) with classifier-free guidance strategy~\cite{ho2022classifier} under the guidance scale of $0.25$, and then fine-tune the model for $4500$ iterations (batch size of 128) using a text recognizer~\cite{retsinas2022best} with a CTC loss~\cite{graves2006connectionist,zhuang2021new,huang2021context}, on four RTX3090 GPUs. The fine-tuning process forces our One-DM to generate readable text with accurate content. The optimizer is AdamW~\cite{loshchilov2017decoupled}, with a learning rate of $10^{-4}$. During inference, each style sample is randomly sampled from the target-writer.
To speed up the sampling, we use the denoising diffusion implicit model (DDIM)~\cite{SongME21} with $50$ steps. More details are provided in~\cref{imp_1}. 


\textbf{Compared methods.} We compare our One-DM with state-of-the-art handwritten text generation methods, including GAN-based methods (i.e., GANwriting~\cite{kang2020ganwriting}, HWT~\cite{bhunia2021handwriting}, VATr~\cite{pippi2023handwritten}, TS-GAN~\cite{DavisMPTWJ20}, HiGAN$+$\cite{gan2022higan+}), and diffusion-based methods (\ie GC-DDPM~\cite{Ding2023ImprovingHO} and WordStylist~\cite{icdar_NikolaidouRCSSSML23}). For a fair comparison, we configure all methods to generate images with a height of 64 pixels, as detailed in~\cref{imp_2}. Additionally, in~\cref{32-pixel}, we compare a variant of One-DM that generates 32-pixel high images with official VATr~\cite{pippi2023handwritten} and HWT~\cite{bhunia2021handwriting}.

\subsection{Main Results}

\textbf{Styled Handwritten Text Generation.} Initially, we assess our One-DM for producing styled handwritten text images, aiming to replicate both the style and content in the generated images. Following~\cite{kang2020ganwriting,bhunia2021handwriting,pippi2023handwritten}, we first calculate FID between generated and real samples for each writer separately and finally average them. In line with the previous works~\cite{kang2020ganwriting,bhunia2021handwriting,pippi2023handwritten}, our experiments on the IAM dataset are divided into four distinct scenarios: IV-S, IV-U, OOV-S, OOV-U. Among these four scenarios, OOV-U represents the most challenging case where both the target style and the words are entirely unseen during training. For the CVL dataset, we directly report the results of all methods on the test set.

\begin{table}[t]
	\caption{Comparisions with state-of-the-art methods on styled and style-agnostic handwritten text generation in the IAM dataset. Note that,  all methods are trained on the same training set used in GANwriting~\cite{kang2020ganwriting}, HWT~\cite{bhunia2021handwriting}, and VATr~\cite{pippi2023handwritten}.
 }
    \vspace{-0.2in}
    \begin{center}
\scalebox{0.75}{
    \begin{threeparttable} 
	\begin{tabular}{lcccccccc}\toprule
      \multirow{2}*{Method} & \multirow{2}*{Shot} &  \multicolumn{4}{c}{Styled Evaluation}  &  \multicolumn{3}{c}{Style-agnostic} \\ \cmidrule(l){3-6}\cmidrule(l){7-9}
      ~  & ~ & \makecell[c]{IV-S}    & \makecell[c]{IV-U}  & \makecell[c]{OOV-S} &   \makecell[c]{OOV-U} & ~~~ & FID$\downarrow$& GS$\downarrow$\\ \midrule
    TS-GAN~\cite{DavisMPTWJ20} & One & 118.56 & 128.75  & 127.11 & 136.67 & & 20.65 &  4.88$\times 10^{-2}$ \\ 
        GANwriting~\cite{kang2020ganwriting} & Few & 120.07  & 124.30  &  125.87 & 130.68 & & 28.37&5.67$\times 10^{-2}$ \\
    HiGAN$+$\cite{gan2022higan+} & One & 117.33  & 116.95  & 121.55 & 121.48  & & 22.95 & 2.06$\times 10^{-2}$ \\
      GC-DDPM~\cite{Ding2023ImprovingHO} & One & 99.86 & 105.73 & 112.52 & 118.39 & & 19.05 &  1.31$\times 10^{-2}$ \\
      WordStylist~\cite{icdar_NikolaidouRCSSSML23} &  One & 98.10 & 104.27 & 109.45 & 115.52  &  & 18.58 & 2.85 $\times 10^{-2}$\\
      HWT~\cite{bhunia2021handwriting} & Few & 109.25  & 106.90 & 116.55 & 113.52 & & 18.99 & 4.41$\times 10^{-3}$ \\
        VATr~\cite{pippi2023handwritten} & Few
        & 103.75 & 101.73   & 111.64 &  108.76 & & 16.03 & 1.74$\times 10^{-2}$  \\
        Ours (One-DM) & One & \textbf{89.47} & \textbf{98.36}   & \textbf{93.30}  & \textbf{102.75}&  & \textbf{15.73} & \textbf{1.98$\mathbf{ \times 10^{-3}}$} \\
        \bottomrule
	\end{tabular} 
    \end{threeparttable}
    }
    \end{center}
    \label{tab:sota-handwritten-metric} 
    \vspace{-0.2in}
\end{table}

We first report the quantitative results on the IAM dataset in~\cref{tab:sota-handwritten-metric}. We can observe that our One-DM outperforms all the competitors in all settings.  Notably, it significantly exceeds one-shot methods in all scenarios. Impressively, our One-DM also holds a substantial advantage over few-shot methods (GANwriting~\cite{kang2020ganwriting}, HWT~\cite{bhunia2021handwriting}, VATr~\cite{pippi2023handwritten}), which use 15-x more reference samples for style guidance, in IV-S and OOV-S settings. Even in the most challenging OOV-U scenario, our One-DM leads the second-best, VATr, by a large margin (102.75 $vs.$ 108.76), demonstrating the superior performance of our One-DM in stylized handwritten text generation. Similarly, our method outperforms HWT and VATr on the CVL dataset, achieving the lowest FID score, as shown in~\cref{cvl}.

\begin{figure*}[t]
    \centering 
    
\includegraphics[width=0.88\linewidth]{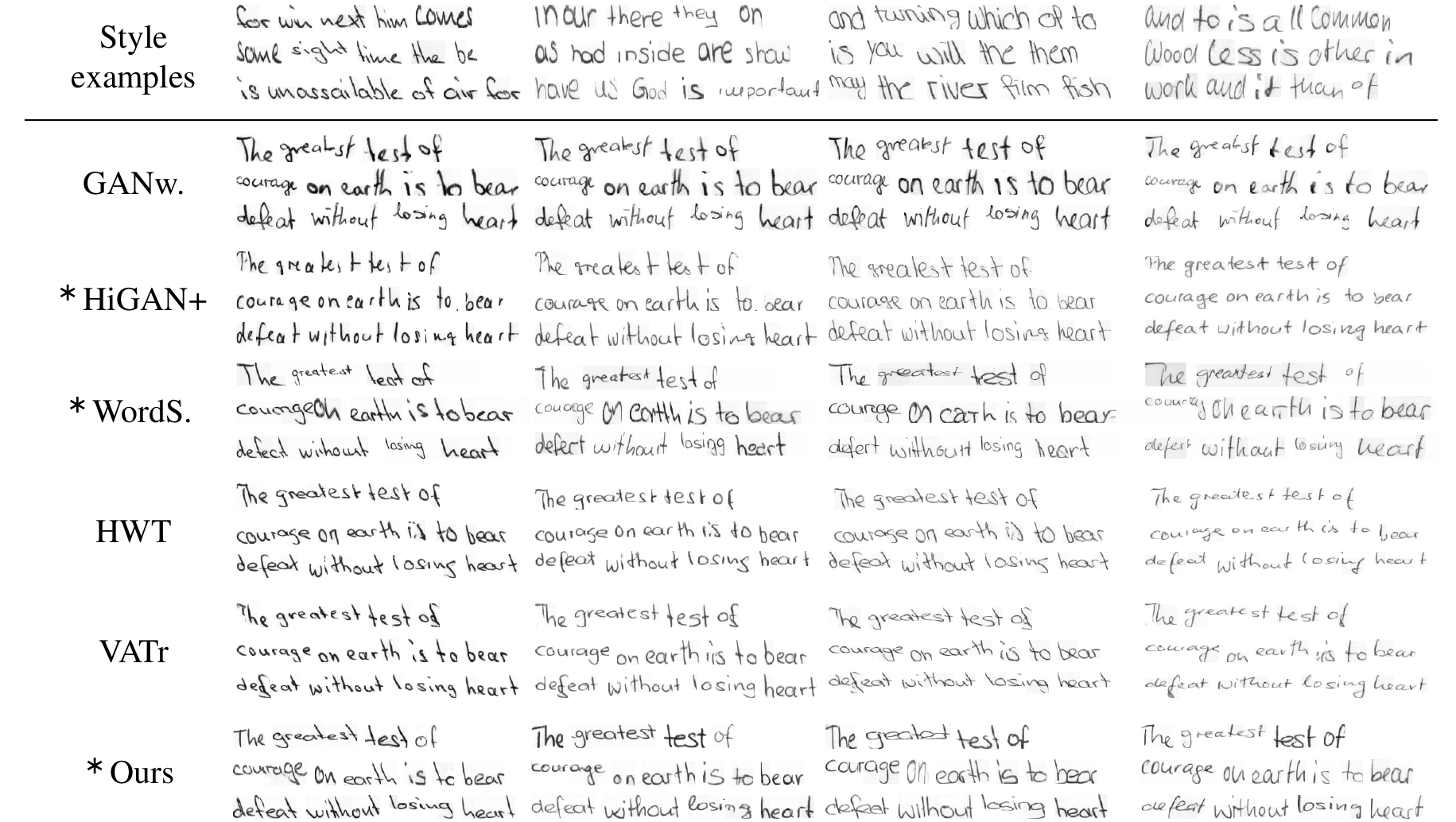}
    \caption{Qualitative comparisons between our method with state-of-the-art methods on handwritten text generation with both specific textual content and desired handwriting style in the IAM dataset. We utilize the identical guiding text, `\textit{The greatest test of courage on earth is to bear defeat without losing heart},' across all handwriting generation methods, directing them to produce text in varied styles. Better zoom in 200\%. $\ast$ denotes one-shot methods, while others are few-shot methods.
    }
    \label{fig:wordvis}
    \vspace{-0.1in}
\end{figure*}

\begin{figure*}[t]
\begin{minipage}{0.48\linewidth}
    \centering \includegraphics[width=0.87\linewidth]{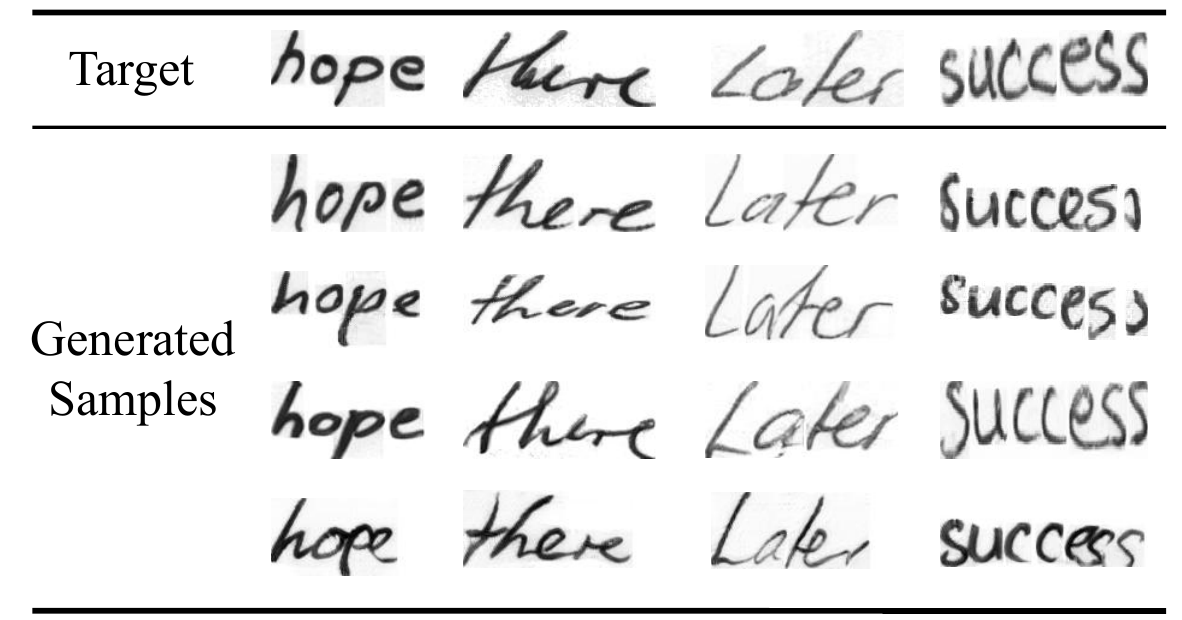}\vspace{-0.1in}
    \caption{Each row shows results from our One-DM and few-shot methods on IAM dataset; readers are invited to identify our method. The answer is at the paper's end.
    } 
\label{fig:guess}
\vspace{-0.2in}
\end{minipage}
\hfill
\begin{minipage}{0.48\linewidth}
   \centering
    \includegraphics[width=0.9\linewidth]{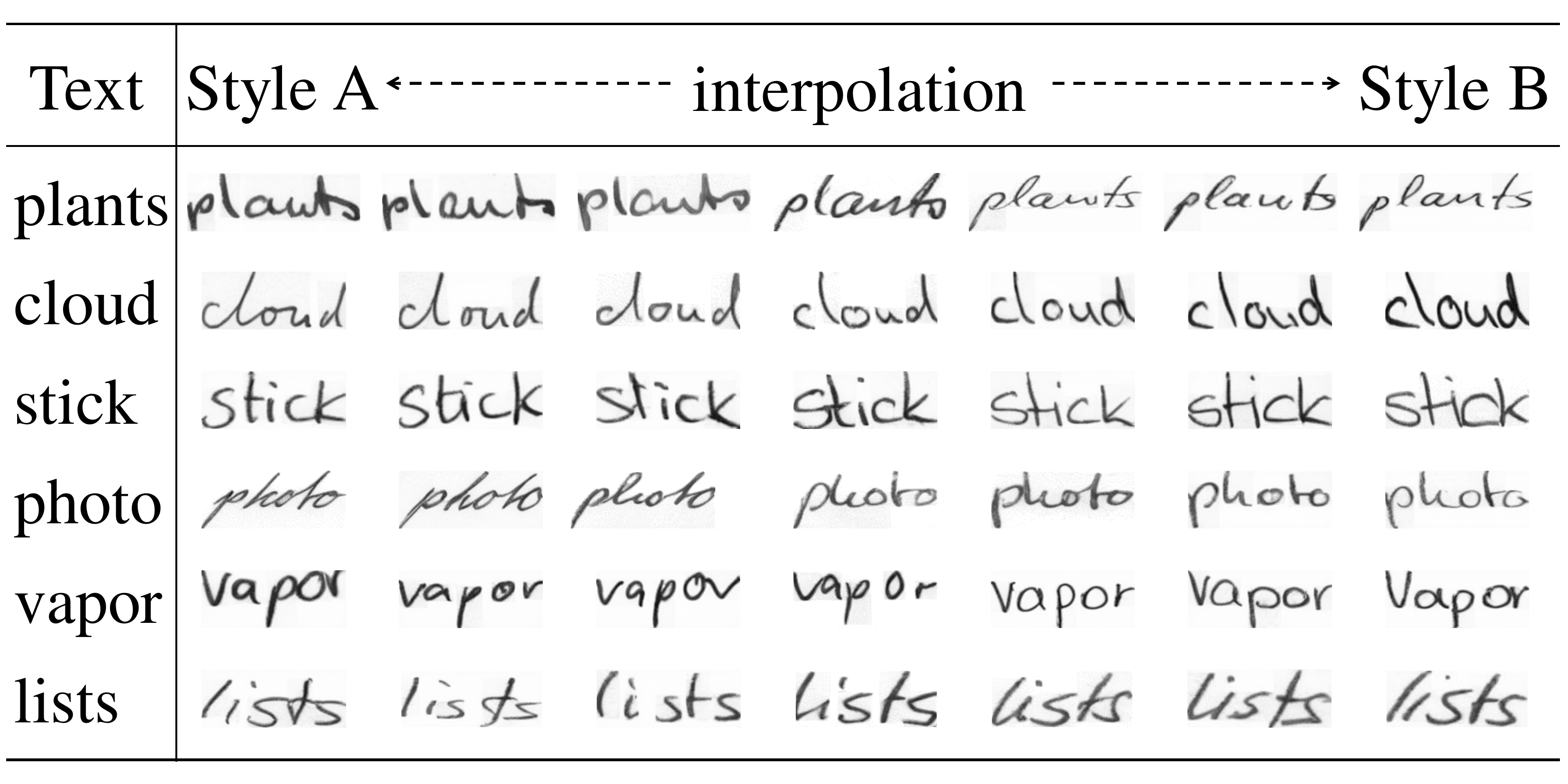}\vspace{-0.1in}
    \caption{We provide style interpolation results generated by our One-DM. Different individual writing styles are extracted from the IAM test dataset.}
    \label{style_interpolation}
\vspace{-0.2in}
\end{minipage}

\end{figure*}

We provide qualitative results to intuitively explain the benefit of our One-DM, in~\cref{fig:wordvis}. GANwriting struggles to capture the style patterns of reference samples, such as character slant and occasionally produces unclear character shapes. HiGAN$+$ more consistently generates characters with correct content, but the character spacing within the generated words lacks realism. WordStylist typically produces images with noticeable background noise. HWT and VATr can produce satisfactory handwritten words in terms of content accuracy and style mimicry; however, their downside is the tendency to create smoother character appearances. Compared to HWT and VATr, our synthesized samples excel in more authentic character ink color and stroke thickness. However, some generated samples by our One-DM are visibly different in ink color. We provide comprehensive explanations in~\cref{ink}. We further display more qualitative comparisons between our method and few-shot methods in~\cref{fig:guess}. 


\textbf{Style-agnostic Handwritten Text Generation.} We further evaluate our One-DM to generate realistic handwritten text images, irrespective of style imitation. For this purpose, we compute the FID and GS on the IAM test set, under the same conditions as applied to ScrabbleGAN~\cite{fogel2020scrabblegan} (FID: 20.72, GS:2.56$\times10^{-2}$), capable of generating handwritten texts with randomly sampled styles. Specifically, each method generates 25k random samples to calculate FID against 25k samples from the test set, and 5k random samples for GS calculation in comparison with 5k test set samples. As presented in~\cref{tab:sota-handwritten-metric}, our One-DM achieves the best results in both FID and GS metrics, further demonstrating its capability to generate higher-quality handwritten text images.

\subsection{Analysis}
In this section, we conduct ablation studies to analyze our One-DM. More analyses are provided in suplementary, including generalization evaluation on different style backgrounds, generation quality assessment through OCR performance, failure case analysis, and the effects of different designs (\eg, high-frequency filter, style-content fusion mechanism, and style input sample length).

\begin{table}[t]
\begin{minipage}{0.51\linewidth}
   \centering
   \vspace{-0.12in}
     \caption{Ablation study. Effect of the Laplacian branch and gate mechanism on the IAM dataset under the OOV-U setting, as done in~\cite{bhunia2021handwriting}.In the middle, we showcase the generated samples of each component.}

    \label{ablation}
    
    \renewcommand\arraystretch{1.78}
    {
    \scalebox{0.8}{
    \begin{tabular}{ccc |cc |c}
     \hline
 \multirow{2}*{Base}   & \multirow{2}*{$\mathcal{E}_{fre}$}   &\multirow{2}*{Gate} & 
         \multicolumn{2}{c|}{Style   samples}
         &  \multirow{2}*{FID $\downarrow$} \\
     \cline{4-5}
     
  ~  &   ~ &  ~ & 
     
     \includegraphics[scale=0.35,valign=c]{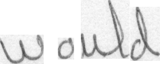} &
        \includegraphics[scale=0.35,valign=c]{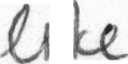}
        & ~   \\
     \hline
        \hline
        

      $\checkmark$&     &    & \includegraphics[scale=0.35,valign=c]{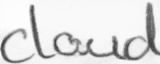}
         & \includegraphics[scale=0.35,valign=c]{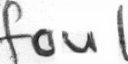}  &  108.44  \\
   $\checkmark$&       & $\checkmark$  & \includegraphics[scale=0.35,valign=c]{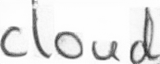}
         & \includegraphics[scale=0.25,valign=c]{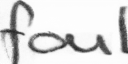} &  105.73
        \\
  $\checkmark$&      $\checkmark$  &    & \includegraphics[scale=0.35,valign=c]{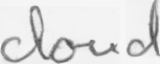}
         & \includegraphics[scale=0.35,valign=c]{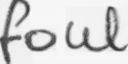} &  104.52
        \\
   $\checkmark$&    $\checkmark$     &  $\checkmark$  & \includegraphics[scale=0.35,valign=c]{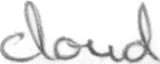}
         & \includegraphics[scale=0.25,valign=c]{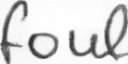} &  102.75  \\
         
        \hline
    \end{tabular}}
    } 
\end{minipage}\hspace{0.2in}
\begin{minipage}{0.38\linewidth}
    \begin{minipage}{1.0\linewidth}

     \caption{Effect of each part of our Laplacian branch on IAM dataset under OOV-U setting.} 
    
    \vspace{-0.25in}
    \begin{center}
\scalebox{0.78}{
    \begin{threeparttable} 
	\begin{tabular}{ccc}\toprule
        $H_s$  &  $\mathcal{L}_{lapNCE}$    & FID $\downarrow$  \\ \midrule
        &  &  108.44 \\
     $\checkmark$  &   & 106.16  \\
        & $\checkmark$  &  107.58 \\
        $\checkmark$  & $\checkmark$  & 104.52  \\
        \bottomrule
	\end{tabular} 
    \end{threeparttable}
    }
    \end{center}
\vspace{-0.05in}
    \label{tab:separated}
\end{minipage}

\begin{minipage}{1.0\linewidth}
  \caption{Comparisons with competitors on the CVL dataset.
    }
   \vspace{-0.25in}
    \centering
    \begin{center}
\scalebox{0.78}{
    \begin{threeparttable} 
	\begin{tabular}{lcc}\toprule
        Method  &  FID $\downarrow$  \\ \midrule
        HWT~\cite{bhunia2021handwriting}&    58.22 \\
     VATr~\cite{pippi2023handwritten}  &    54.44  \\
      Ours (One-DM)  &  51.78 \\
        \bottomrule
	\end{tabular} 
    \end{threeparttable}
    }
    \end{center}
    \label{cvl}

\end{minipage}
\end{minipage}
\vspace{-0.2in}
\end{table}

\textbf{Quantitative evaluation of Laplacian branch and gate mechanism.}
We conduct various ablation experiments on the IAM dataset to evaluate the effect of distinct components within our approach. We provide the quantitative result in ~\cref{ablation}. We find that: (1) The inclusion of both Laplacian branch and gate mechanism enhances the quality of the generated handwritten text images, improving FID by 3.92 and 2.71, respectively. (2) Integrating the Laplacian branch with the gate mechanism further boosts the generative performance. 

\textbf{Qualitative evaluation of Laplacian branch and gate mechanism.}
To further analyze each module in our One-DM, we conduct visual ablation experiments. As shown in~\cref{ablation}. we can observe that firstly, 
after adding a gate mechanism, background noise can be somewhat suppressed, resulting in characters with relatively clean backgrounds. Then, the independent addition of the Laplacian branch helps the model learn cursive connections and other style patterns. Finally, our method integrated the Laplacian branch and gate mechanism, which can generate the highest quality handwritten text images.

\textbf{Discussions of the Laplacian branch.} Our Laplacican branch consists of two key components: utilizing high-frequency images $H_s$ and the Laplacian contrastive learning loss $\mathcal{L}_{lapNCE}$. In the previous ablation study, they are always combined. We further conduct exploration experiments to explain why they cannot be separated. As reported  in~\cref{tab:separated}, combining $H_s$ and  $\mathcal{L}_{lapNCE}$ maximizes effectiveness; separating them significantly reduces performance.
Without the guidance of $\mathcal{L}_{lapNCE}$, extracting discriminative features from $H_s$  is challenging. Likewise, directly applying $\mathcal{L}_{lapNCE}$ on original images leads to an undesired style feature extraction, as original images have less clear style patterns than $H_s$.

\textbf{Discussions about learning style from a single reference.} We are quite surprised that One-DM, with just a single reference sample, even exceeds the generative performance of few-shot methods. We provide the potential reason analysis below. Firstly, One-DM learns a meaningful style latent space, wherein new styles can be generated based on seen styles (cf.~\cref{style_interpolation}). Then, through our style-enhanced module, One-DM effectively extracts the style feature from a single example and maps it to a position close to the exemplary writer in the feature space, thereby producing high-quality styled handwritten text images.

\begin{figure*}[t]
    \centering \includegraphics[width=0.95\linewidth]{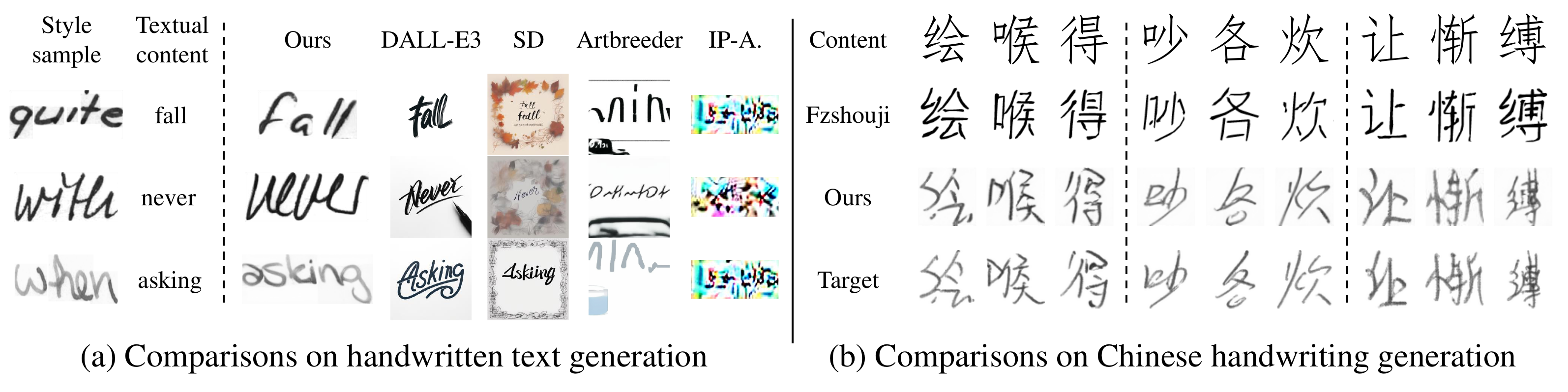}
\vspace{-0.1in}
    \caption{(a) Qualitative comparisons between our method with SOTA industrial image generation methods, including DALL-E3~\cite{betker2023improving}, Stable Diffusion~\cite{rombach2022high}, Artbreeder\textsuperscript{\ref {Artb}} and IP-Adapter\textsuperscript{\ref {IPA}} on handwritten text generation with both controllable styles and contents. 
    (b) Qualitative comparisons with Fzshouji\textsuperscript{\ref{shouji}} on ICDAR-
2013 competition database~\cite{yin2013icdar}.
    }
   \vspace{-0.1in}
\label{fig:industrial}

\end{figure*}

\subsection{Comparisons with SOTA Industrial Methods}
To highlight the superiority of our method, we compare One-DM with leading industrial image generation methods that are trained on tremendously large datasets (including numerous text-centric images), including two prominent text-to-image methods, DALL-E3~\cite{betker2023improving} and Stable Diffusion~\cite{rombach2022high} (SD), and two popular style transfer methods, Artbreeder\footnote{\url{https://www.artbreeder.com/}\label{Artb}} and IP-Adapter\footnote{\url{https://github.com/tencent-ailab/IP-Adapter}\label{IPA}} (IP-A.), on the IAM dataset. Further experiment details are in~\cref{ind}. 

As shown in \cref{fig:industrial}(a), our method excels industrial methods in style mimicry and content preservation. The performance of IP-A. is the poorest, often producing distorted images. Artbreeder can replicate the color of the strokes from style samples, but fails in content preservation. DALL-E3 and SD generate characters with accurate content, but often mismatch style details with references, such as character spacing and stroke width, with SD often generating extra backgrounds. Besides, we compare Fzshouji\footnote{\url{https://www.fzshouji.com/make}\label{shouji}}, an advanced  industrial method designed for Chinese handwriting generation. As shown in~\cref{fig:industrial}(b), our method outperforms Fzshouji in replicating character details and ink color.

\begin{table}[t]
\begin{minipage}{0.46\linewidth}
 \caption{{Quantitative comparisons with competitors on styled handwritten character generation in terms of FID.}
    }
    \vspace{-0.2in}
    \begin{center}
\scalebox{0.8}{
    \begin{threeparttable} 
	\begin{tabular}{lccc}\toprule
 Method   &  Chinese &   & Japanese \\ \midrule GANwriting~\cite{kang2020ganwriting} &    116.49 & & 111.86 \\
  HWT~\cite{bhunia2021handwriting} &     165.74 &  & 148.66 \\
    VATr~\cite{pippi2023handwritten} &   139.91 & & 124.98 \\
    WordStylist~\cite{icdar_NikolaidouRCSSSML23}   &34.61 &  & 101.93 \\
    Ours (One-DM) &  27.24 &  & 95.43 \\
        \bottomrule
	\end{tabular} 
    \end{threeparttable}
    }
    \end{center}
    \label{tab:character}
    
\end{minipage}
\hfill
\begin{minipage}{0.47\linewidth}
  \centering \vspace{-0.1in}
    \caption{Qualitative comparisons with VATr~\cite{pippi2023handwritten} on Chinese dataset.}
    \vspace{-0.05in}
    \includegraphics[width=0.85
    \linewidth]{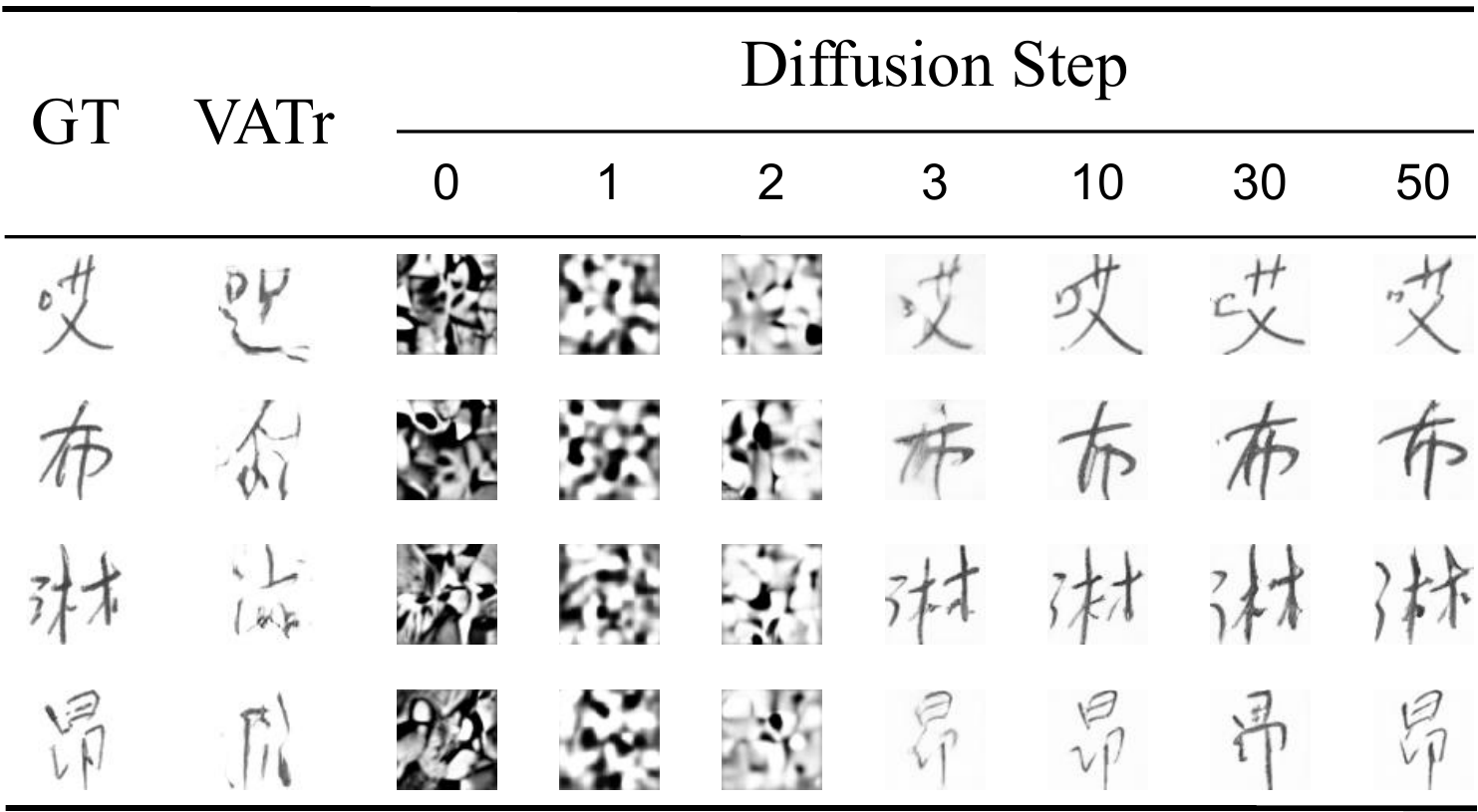}
    
    \label{sampling_timestep}

\end{minipage}
\vspace{-0.3in}
\end{table}

\subsection{Applications to Other Languages}

\begin{figure*}[t]
    \centering \includegraphics[width=0.78\linewidth]{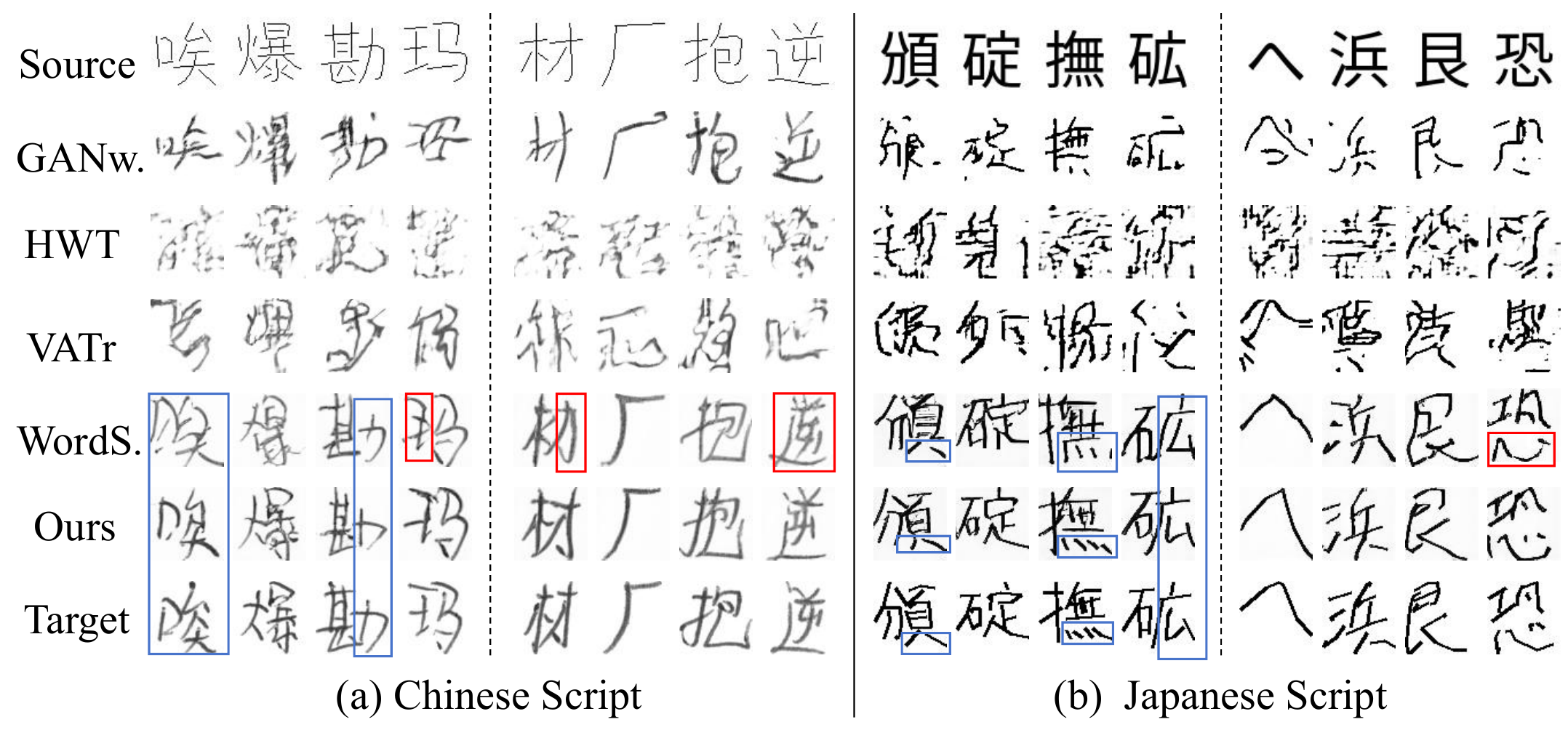}\vspace{-0.1in}
    \caption{Comparisons with GANwriting~\cite{kang2020ganwriting}, HWT~\cite{bhunia2021handwriting}, VATr~\cite{pippi2023handwritten}, and WordStylist~\cite{icdar_NikolaidouRCSSSML23} for styled handwritten character generation on Chinese and Japanese scripts. The red boxes highlight failures of structure preservation, while blue boxes highlight comparisons between the style patterns of targets and generated characters.
    }
\label{fig:Chinese}
\vspace{-0.25in}
\end{figure*}
In this section, we evaluate whether One-DM can be used to generate languages other than English. We further conduct experiments in the Chinese (\ie ICDAR-2013 competition database~\cite{yin2013icdar}) and Japanese(\ie UP\_Kuchibue database~\cite{jaeger2001two}) datasets.  We use FID to assess generated samples of each writer, then average them. More experiment details can be seen in~\cref{app}.


For the Chinese handwritten character generation task, as shown in~\cref{tab:character}, we find that our One-DM outperforms the second-best method by a large margin, achieving a $7.37$ lower FID.
From \cref{fig:Chinese} (a), we can observe that 
our One-DM generates characters that closely match the target images in geometric shape and character slant. In contrast, handwritings from HWT and VATr exhibit noticeable artifacts like blurring and collapsed structures. GANwriting tends to miss strokes in its handwritings. WordStylist sometimes struggles to accurately mimic style patterns and tends to generate characters with incorrect radicals. \cref{tab:character} and \cref{fig:Chinese} (b) further verify the effectiveness of One-DM for Japanese handwriting generation. We also achieve the lowest FID score, and our generated Japanese samples excel in both content preservation and style imitation.

We further investigate why diffusion-based methods (Our One-DM and WordStylist) that require only one single sample outperform few-shot GAN-based methods (i.e., GANwriting, HWT, and VATr) so significantly in generating Chinese and Japanese characters. 
The lower performance of GAN-based methods on Chinese and Japanese characters may stem from their vanilla convolutional architectures struggling with Chinese and Japanese characters' complex geometry, as noted in~\cite{xie2021dg}. In contrast, our One-DM breaks down the generation of Chinese and Japanese characters into simpler steps. For instance, as demonstrated in ~\cref{sampling_timestep}, during the early stages of the diffusion generation process, the model first attempts to generate a rough Chinese handwritten character. It then continues to refine the writing style (e.g., character shape and stroke color) under the condition guidance, until it synthesizes handwritings that are satisfactory. 


\section{Conclusion}
In this paper, we introduce a novel One-DM for handwritten text generation, requiring only one style reference to produce realistic handwritten text images\footnote{In~\cref{fig:guess}, the second from bottom row belongs to our method.}. We enhance style extraction by incorporating high-frequency components from the style reference. For high-frequency components with distinct style patterns, we employ laplacian contrastive learning to capture more discriminative style features. Moreover, a gate mechanism improves the transfer of informative features from the reference, reducing background noise. Our One-DM outperforms few-shot methods in multiple language scripts. In the future, we aim to explore the potential of One-DM in font generation and vector font creation tasks. 

\noindent\textbf{Acknowledgments}
The research is partially supported by National Key Research and Development Program of China (2023YFC3502900), National Natural Science Foundation of China (No.62176093, 61673182), Key Realm Research and Development Program of Guangzhou (No.202206030001), Guangdong-Hong Kong-Macao Joint Innovation Project (No.2023A0505030016).

%
%

\bibliographystyle{splncs04}
\bibliography{main}

\clearpage
\begin{spacing}{1.5}
{\centering
\textbf{\Large{Supplementary Material: \\ One-DM: One-Shot Diffusion Mimicker for Handwritten Text Generation}}
\par}
\end{spacing}

\setcounter{equation}{0}
\renewcommand\theequation{\alph{equation}}

\setcounter{section}{0}
\renewcommand\thesection{\Alph{section}}
\label{sec:rationale}

We organize our supplementary material as follows.
\begin{itemize} 
\item In~\cref{related}, we discuss several similar works, and review the related work of diffusion methods in general image generation. 

\item In~\cref{diffusion}, we describe more method details about the conditional diffusion model of our One-shot Diffusion Mimcker (One-DM).
\item In~\cref{imp}, we provide more implementation details.
\item In~\cref{32-pixel}, we report additional quantitative results, comparing our One-DM with official VATr and HWT for generating 32-pixel high images.


\item In~\cref{user_study}, we provide user study experiments.

\item In~\cref{stylebg}, we explore the generalization of our One-DM to style images with various noisy backgrounds.

\item In~\cref{ocr}, we assess the generation quality through OCR system.

\item In~\cref{fail}, we conduct failure case analysis.

\item In~\cref{ablation-supp}, we analyze the impact of different high-frequency filters, style-content fusion mechanisms, and style sample lengths.

\item  In~\cref{ink}, we provide more analysis experiments to explain the different ink color in Figure \textcolor{red}{4} of Section \textcolor{red}{4.3}.

\item In~\cref{ADD_SCR}, we show more visual comparisons of handwritten text generation, covering English, Chinese, and Japanese scripts.
\end{itemize}

\section{More Related Work}\label{related}

\textbf{Discussion on several similar works.} Our Laplacian contrastive learning $\mathcal{L}_{lapNCE}$ differs from the CC loss~\cite{xie2022toward} in one key aspect: $\mathcal{L}_{lapNCE}$ is applied to \emph{high-frequency} information $H_s$ for enhancing \emph{handwriting style features}, whereas the CC loss is applied to text images for extracting \emph{scene text content features}. Ours is a specialized design for style extraction, where we find simply using $\mathcal{L}_{lapNCE}$ on images leads to poor performance (cf.~Table \textcolor{red}{3} in Section~\textcolor{red}{4.3}). The idea of using $H_s$ also differs from prior work~\cite{lin2021drafting,pan2023visa}. We aim to enhance handwriting style extraction, whereas~\cite{lin2021drafting} targets improving content representations of general images, and~\cite{pan2023visa} aims to address domain shifts of scene texts.

\textbf{Image diffsuion.} The Denoising Diffusion Probabilistic Model (DDPM)~\cite{ho2020denoising}, a typical representative of diffusion models, has gained extensive application in image generation task~\cite{nichol2021glide,saharia2022photorealistic,ruiz2023dreambooth,zhang2023adding,zhang2023hipa,liu2023geom,cao2019multi,huang2022agtgan,liang2024diffusion,xu2024kmtalkspeech}. To reduce computational expenses, latent diffusion model (LDM)~\cite{rombach2022high} is introduced, executing diffusion steps in latent space rather than in pixel space, as done in DDPM. Furthermore, guided diffusion~\cite{dhariwal2021diffusion} incorporates an additional classifier to control the categories of synthesized images. After that, classifier-free diffusion model~\cite{ho2022classifier} explores the conditional image synthesis task without the separate classifier guidance. Recently, some text-guided image generation methods have been proposed, such as DALL-E3~\cite{betker2023improving}, Stable Diffusion~\cite{rombach2022high}, GLIDE~\cite{nichol2021glide}, and Imagen~\cite{saharia2022photorealistic}. These methods employ pre-trained text encoders, such as CLIP~\cite{radford2021learning}  and BERT~\cite{devlin2019bert}, to convert given text descriptions into comprehensive representations, thereby guiding the diffusion process to produce realistic and diverse images. 

Different from general text-to-image generation tasks, the users' handwriting style, as an abstract concept in nature, presents considerable challenges in its accurate depiction through text descriptions. In response to this, we introduce an innovative style-enhanced module that extracts discriminative style representations from individual handwritings, which are then used as guiding conditions for the diffusion model.

\section{More Details of the Conditional Diffusion Model}\label{diffusion}
As described in Section \textcolor{red}{3.4}, the training objective of our conditional diffusion model is:
\begin{equation} \label{loss}
\mathcal{L}_{rec}\textbf{\small{=}}\mathbb{E}_{t,q} \left\| \mu_t(x_t,x_0)-\mu_{\theta}(x_t,g,t)\right\|_2^2.
\end{equation}
The derivation formula for $\mu_t(x_t,x_0)$, provided by Denoising Diffusion Probabilistic Models (DDPM)~\cite{ho2020denoising}, is given by:
\begin{equation} \label{gt}
\mu_t(x_t,x_0)\textbf{\small{=}}\frac{\sqrt{\alpha_t}(1-\overline{\alpha}_{t-1})x_t+\sqrt{\overline{\alpha}_{t-1}}(1-\alpha_t)x_0}{1-\overline{\alpha}_t},
\end{equation} 
where $\alpha_t\text{\small{=}}1-\beta_t$, $\beta_t$ is a variance schedule, and $\overline{\alpha}_t\text{\small{=}}\prod \limits_{i=1}^t\alpha_i$. Recall that ${\mu_{\theta}(x_t,g,t)}$ also depends on $x_t$, we can closely align $\mu_t(x_t,x_0)$ by configuring it in the following manner:
\begin{equation} \label{fake}
\mu_{\theta} (x_t,g,t)\text{\small{=}}\frac{\sqrt{\alpha_t}(1-\overline{\alpha}_{t-1})x_t+\sqrt{\overline{\alpha}_{t-1}}(1-\alpha_t)\hat{x}_{\theta}(x_t,g,t)}{1-\overline{\alpha}_t},
\end{equation} where $\hat{x}_{\theta}(x_t,g,t)$ is implemented by a U-Net~\cite{ronneberger2015u} network designed to estimate $x_0$ from the noisy image $x_t$ conditioned on the guidance $g$ at the time step $t$. Then, \cref{loss}, when \cref{gt} and \cref{fake} are substituted into it, yields the following derivation:

\begin{align}    \label{longeq}
&\mathbb{E}_{t,q} \left\| \mu_t(x_t,x_0)-\mu_{\theta}(x_t,g,t)\right\|_2^2  \nonumber    \\ \nonumber
&=\mathbb{E}_{t,q}\Big\| \frac{\sqrt{\alpha_t}(1-\overline{\alpha}_{t-1})x_t+\sqrt{\overline{\alpha}_{t-1}}(1-\alpha_t)x_0}{1-\overline{\alpha}_t}    \\ \nonumber 
& - \frac{\sqrt{\alpha_t}(1-\overline{\alpha}_{t-1})x_t+\sqrt{\overline{\alpha}_{t-1}}(1-\alpha_t)\hat{x}_{\theta}(x_t,g,t)}{1-\overline{\alpha}_t} \Big\|_2^2     \\ \nonumber
&=\mathbb{E}_{t,q} \left\| \frac{\sqrt{\overline\alpha_{t-1}}(1-{\alpha}_t)}{1-\overline{\alpha}_t}(x_0 - \hat{x}_{\theta}(x_t,g,t)) \right\|_2^2  \\  
&=\mathbb{E}_{t,q} \left[ \frac{\overline\alpha_{t-1}(1-\alpha_t)^2}{(1-\overline\alpha_t)^2}\left\|x_0 - \hat{x}_{\theta}(x_t,g,t)  \right\|_2^2 \right] .
\end{align}
In practice, DDPM observes enhanced training performance of the diffusion model when utilizing a simplified objective that omits the weighting term in \cref{longeq}:
\begin{align}
\mathcal{L}_{rec}=\mathbb{E}_{t,q}  \left\|x_0 - \hat{x}_{\theta}(x_t,g,t)  \right\|_2^2  .
\end{align}
From this equation, we know that optimizing $\mathcal{L}_{rec}$ essentially involves training a U-Net $\hat{x}_{\theta}$ to reconstruct the original image $x_0$ from its noise-altered version $x_t$.

After training, during the diffusion generation process, our conditional diffusion model $p_{\theta}(x_{t-1}|x_t,g)$ progressively generates the desired handwritten text image conditioned on the guidance $g$, following Eq. (\textcolor{red}{6}):
\begin{align}
x_{t-1} = \mu_{\theta} (x_t,g,t) + \sqrt{\Sigma_{\theta}(x_t,g,t)}\epsilon,
\end{align}
where we set $\Sigma_{\theta}(x_t,g,t)\text{\small{=}}\beta_t$ to untrained time-dependent constants and noise $\epsilon$ is randomly sampled from a pure normal distribution $\mathcal{N}(0,\mathbf{I})$, as done in DDPM.

\section{More Experimental Details}\label{imp}
\subsection{More Implementation Details of Our One-DM}\label{imp_1}
Our approach incorporates three encoders: a content encoder featuring two standard transformer layers, and both the high-frequency style encoder and the spatial style encoder, each equipped with three standard transformer layers. Following the original Transformer architecture~\cite{vaswani2017attention,dai2023disentangling,liao2024pptser}, each Transformer layer contains the multi-head attention with c = 512 dimensional states and 8 attention heads. Similarly, in our style-content fusion module, the feature dimension of all the attention modules is set to 512, with 8 heads for each module. Moreover, we apply sinusoidal positional encoding~\cite{vaswani2017attention} to input tokens before feeding them to the Transformer encoder. 

We adopt latent diffusion model (LDM)~\cite{rombach2022high} to implement our conditional diffusion model $p_{\theta}$, which adopts a U-Net architecture~\cite{ronneberger2015u} integrated with transformer blocks to blend noise with the condition $g$. To conserve GPU memory and accelerate the training time, we streamline the U-Net by reducing the number of ResNet blocks and attention heads, as done in~\cite{icdar_NikolaidouRCSSSML23}. LDM takes the diffusion process into the latent space. Specifically, the LDM employs an autoencoder where an encoder  $\mathcal{E}$ encoders the input image $x$ into a lower dimensional latent representation $z\text{\small{=}}\mathcal{E}(x)$ while a decoder $\mathcal{D}$ converts the latent code back to the image space such that perceptually $\widetilde{x}\text{\small{=}}\mathcal{D}(z)\text{\small{=}}\mathcal{D}(\mathcal{E}(x))$. Significantly, the encoder downsamples the image $x \in \mathbb{R}^{H \times W \times 3}$ to obtain the latent code $z \in \mathbb{R}^{H/8  \times W/8 \times4 }$ by a factor $f\text{\small{=}}8$. Thus, the U-Net $\hat{x}_{\theta}$ directly operates on a more compact representation, greatly saving runtime and computational cost. Regarding the specific implementation of the autoencoder, we adopt a powerful, pre-trained Variational Autoencoder (VAE)\footnote{\url{https://huggingface.co/CompVis/stable-diffusion}}. During the training phase, we freeze the parameters of VAE and we set $T=1000$ steps, and forward process variances are set to constants increasing linearly from $\beta_{1}=10^{-4}$ to $\beta_{T} = 0.02$.

\subsection{More Implementation Details of Compared Methods}\label{imp_2}
We use official implementations provided by authors for methods like GANwriting~\cite{kang2020ganwriting}, TS-GAN~\cite{DavisMPTWJ20}, and HiGAN$+$\cite{gan2022higan+} that are inherently capable of generating 64-pixel high images. For VATr~\cite{pippi2023handwritten} and HWT~\cite{bhunia2021handwriting}, which produce 32-pixel high images, we simply modify their code by adding a convolution layer followed by an upsampling to enable 64-pixel image generation. To enable WordStylist~\cite{icdar_NikolaidouRCSSSML23} and GC-DDPM~\cite{Ding2023ImprovingHO} to extract style patterns from one single style sample, we replace their writer-ID embedding layers with our spatial style encoder. We update the WordStylist~\cite{icdar_NikolaidouRCSSSML23} based on its official code and re-implement the GC-DDPM~\cite{Ding2023ImprovingHO}; both inherently generate images with a height of 64 pixels.

\subsection{More Details of the Comparisons with Industrial Methods}\label{ind}
We compare One-DM with text-to-image methods, DALL-E3~\cite{betker2023improving} and Stable Diffusion~\cite{rombach2022high} (SD), and style transfer methods, Artbreeder\footnote{\url{https://www.artbreeder.com/}\label{Art}} and IP-Adapter\footnote{\url{https://github.com/tencent-ailab/IP- Adapter}\label{IP}} (IP-A.), as mentioned in Section \textcolor{red}{4.4}. To ensure a fair comparison, for text-to-image models that require the text descriptions of style samples, we initially feed sample images into GPT-4\cite{achiam2023gpt} to comprehend the handwriting style of the input samples,  subsequently generating corresponding style prompts, such as ``Tilted to the right, small curvature, very thin strokes, black color, some hyphenation''. After that, with the obtained style prompts and specific text content, we guide the text-to-image models to generate desired handwritten text images. For instance, if we want to generate a styled handwritten `great', we feed the generated style prompts and the instruction prompt ``Please generate a handwriting image with the text `great' based on the above style description on white paper without any background decoration!'' into the text-to-image models. As for the style transfer methods, they can directly handle the style and content images without any extra adaptation. 

\subsection{More Details on  Applications to Other Languages}\label{app}
\textbf{Dataset.} For the Chinese handwritten character generation task, we use the ICDAR-2013 competition database~\cite{yin2013icdar}, which contains 60 writers and 3755 different Chinese characters for each writer. We randomly select $80\%$ of the entire dataset as the training set, and the remaining $20\%$ as the test set. As for source content images, we use the popular average Chinese font~\cite{jiang2019scfont}. For the Japanese handwriting generation task, we conduct experiments on the UP\_Kuchibue database~\cite{jaeger2001two} to evaluate the effectiveness of our method. This database contains about 340k online Japanese characters belonging to 103 writers. We render the offline character images by connecting coordinate points of online characters. For the source content images, we employ printed Japanese font. Similarly, we allocate $80\%$ of the entire dataset randomly for training and reserve the remaining $20\%$ for testing. In the experiments, we resize all images to $64\times64$.

\textbf{Compared Methods.} Since our One-DM can imitate any handwriting style and synthesize handwritten text with arbitrary content, it's readily applicable to various handwriting tasks without requiring adjustments. We compare our One-DM with state-of-the-art few-shot handwritten text generation methods, i.e., GANwriting~\cite{kang2020ganwriting}, HWT~\cite{bhunia2021handwriting} and VATr~\cite{pippi2023handwritten} and diffusion-based method, i.e. WordStylist~\cite{icdar_NikolaidouRCSSSML23}. Our method and WordStylist require only one randomly sampled style sample as the style input, whereas other few-shot methods continue to use 15 samples. For all competitors, we follow their official experiment hyper-parameters to ensure they are all thoroughly trained.

\begin{table}
	\caption{Additional comparisons with official HWT and VATr on stylized handwritten text generation in terms of the FID score. The best results are in bold.}
 \label{add_fid}
    \vspace{-0.2in}
    \begin{center}
\scalebox{1.00}{
    \begin{threeparttable} 
	\begin{tabular}{lcccc}\toprule
        Method  &  \makecell[c]{IV-S}    & \makecell[c]{IV-U}  & \makecell[c]{OOV-S} &   \makecell[c]{OOV-U} \\ \midrule

        HWT~\cite{bhunia2021handwriting}& 106.97  & 108.84 & 109.45 & 114.10  \\
        VATr~\cite{pippi2023handwritten}
        & 88.20 & 91.11   & 98.57 &  102.22 \\
        Ours (One-DM) & \textbf{79.60} & \textbf{86.09}   & \textbf{85.03}  & \textbf{90.84} \\
        \bottomrule
	\end{tabular} 
    \end{threeparttable}
    }
    \end{center}
    \label{tab:sota-handwritten} 
    \vspace{-0.1in}
\end{table}

\section{Additional Comparisons with Official VATr and HWT} \label{32-pixel}
To facilitate a more equitable comparison, we conduct experiments on the IAM dataset with a height of 32 pixels, as provided by VATr~\cite{pippi2023handwritten} and HWT~\cite{bhunia2021handwriting} in their official repositories. Benefiting from our U-Net architecture, the size of the output images remains consistent with the input images, allowing our model to generate images with a height of 32 pixels without any modifications. We train our model with the same training data as used in VATr and HWT. For evaluation, we resize the generated images to a height of 64 while maintaining the aspect ratio of width, as suggested in VATr. We report the results of all methods across four settings, as shown in~\cref{add_fid}. From these quantitative results, we can observe that our One-DM significantly outperforms VATr and HWT across all four settings, further substantiating the superiority of our approach in synthesizing stylized handwritten text.

\begin{figure}
    \centering
\includegraphics[width=0.75\linewidth]{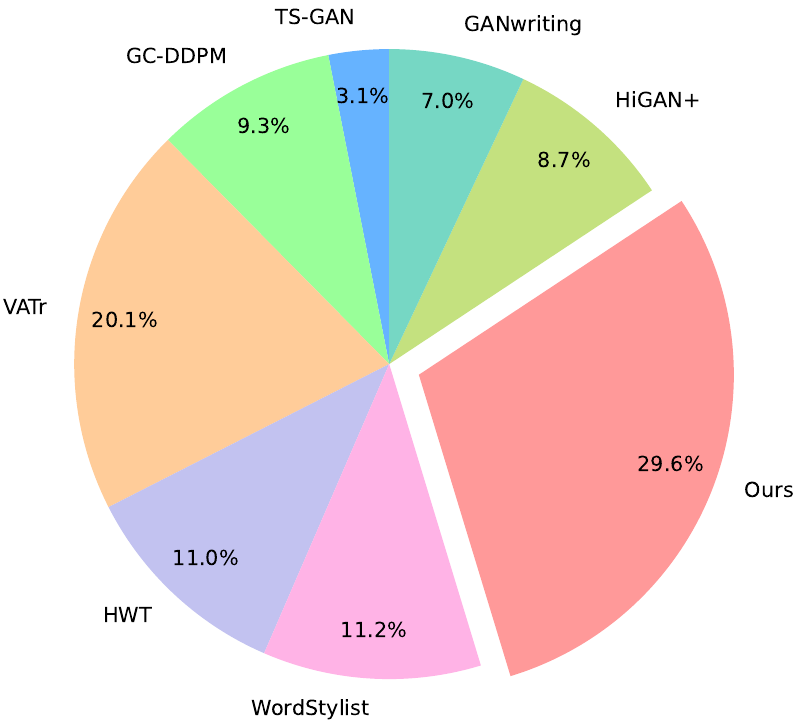}
    \caption{User preference study with a comparison to state-of-the-art methods on handwritten text generation.}
    \label{user_prefer}
    \vspace{-0.1in}
\end{figure}


    

\section{User Studies} \label{user_study}
\textbf{User preference study.} We first invite human participants with a college undergraduate education to evaluate the visual quality of synthesized handwritten text images in terms of style imitation, with the generated samples being from our method and state-of-the-art methods. 
At each time, we randomly select a writer from the IAM dataset, using their handwritten samples as style guidance, and employ identical text as content guidance, to direct all methods in generating candidate samples. Then, given three style references from the exemplar writer along with several candidates generated by different methods, participants are required to pick up the most similar candidate in style with the reference samples. We repeat this procedure 30 times and finally collect 900
valid responses contributed by 30 volunteers. As shown in~\cref{user_prefer}, our method garners the most user preferences, demonstrating that our approach produces the highest quality samples in terms of style mimicry.

\begin{table}
    \centering
    \caption{Confusion matrix$\left(\%\right)$ in user plausibility study. A classification accuracy of $49.05\%$ implies that our synthesized handwritten text images are difficult to distinguish from real images by human users.} 

    \label{plausi}

    \scalebox{1.00}{
    \begin{tabular}{l cc c}
     \toprule

      \multirow{2}*{Actual}   & 
         \multicolumn{2}{c}{Predicted}
         &  \multirow{2}*{\makecell{Classification \\ Accuracy }} \\
         \cmidrule{2-3}
     
      ~ &  Real & Fake    & ~   \\
        \midrule
        
        Real  &  30.22  & 19.78   &  \multirow{2}*{49.05}  \\
        Fake & 31.17 & 18.83 & ~ \\
        \bottomrule
    \end{tabular}}
\end{table}

\textbf{User plausibility study.} We further study whether the generated images from our One-DM are indistinguishable from the real handwriting samples by conducting a user plausibility study. In this experiment, we initially present participants with 30 exemplary real handwritten samples. Subsequently, they are tasked with determining whether each given image is an authentic handwriting or a synthetic one, where each image is randomly drawn both from genuine training images and those synthesized by our method. We collect a total of 1800 valid responses from 30 participants. We report the results as a confusion matrix in~\cref{plausi}, We find that the classification accuracy is close to  $50\%$, indicating that it turned into a random binary classification. This reveals that the images synthesized by our method are nearly perceived as plausible.

\begin{figure}[t]
  \centering
  \includegraphics[width=0.95\linewidth]{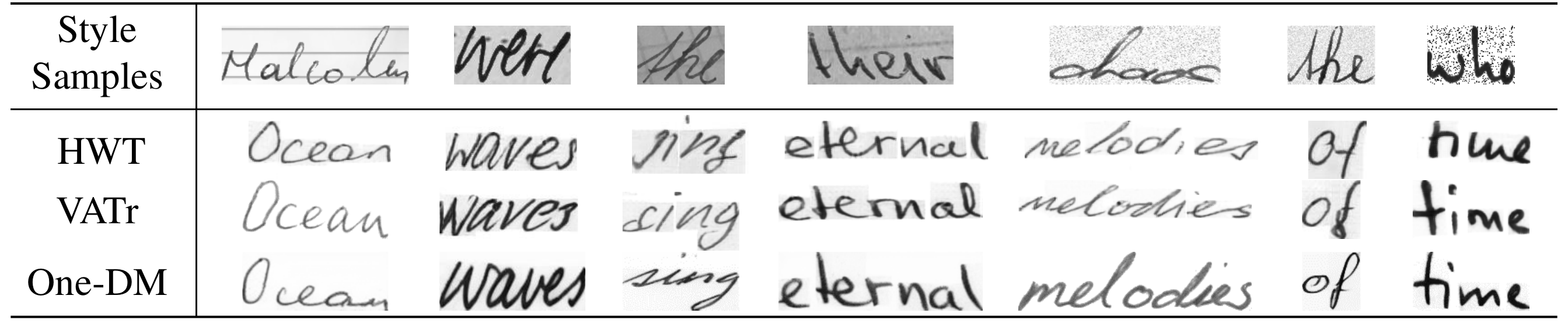}
   \caption{Generated results conditioned on dirty style samples.}
   \label{fig:style-background}
   \vspace{-0.2in}
\end{figure}

\section{Generalization to different style backgrounds}\label{stylebg}
To test the tolerance of our One-DM to background artifacts, we use images with various noisy backgrounds as style references. From ~\cref{fig:style-background}, we find that our one-DM still synthesizes high-quality handwritten text images under style references with dirty backgrounds, demonstrating the robustness of our style extraction to various style backgrounds.

\section{Evaluation of OCR System}\label{ocr}
To further assess the generation quality, we use the generated training sets by our One-DM and VATr to train an OCR system. As shown in~\cref{fig:ocr}, the recognizer~\cite{retsinas2022best} trained on our synthesized dataset outperforms VATr~\cite{pippi2023handwritten} by achieving lower character error rates (CER) and word error rates (WER) on both the real IAM test set and the CVL test set. Notably, our CER and WER performance on the CVL test set even approaches that of the recognizer trained on the real dataset.

\begin{table}[h]
\centering
\caption{OCR performance on IAM and CVL test sets.}
\scalebox{0.9}{
\begin{tabular}{@{}lcccc@{}}
\toprule
 \multirow{2}*{Training data}& \multicolumn{2}{c}{IAM} & \multicolumn{2}{c}{CVL} \\
\cmidrule(r){2-3} \cmidrule(l){4-5}
 & CER$\downarrow$(\%) & WER$\downarrow$(\%) & CER$\downarrow$(\%) & WER$\downarrow$(\%) \\
\midrule[0.75pt]
Real dataset & 4.98 & 13.52 & 16.85 & 22.81 \\
\midrule 
VATr & 26.98 & 60.43 & 30.84 & 49.59 \\
Ours(One-DM) & \textbf{11.69} & \textbf{25.55} & \textbf{17.03} & \textbf{23.52} \\
\bottomrule
\end{tabular}
\label{fig:ocr}
}
\vspace{-0.2in}
\end{table}

\section{Analysis of Failure Cases}\label{fail}
\begin{figure}
    \centering
\includegraphics[width=0.8\linewidth]{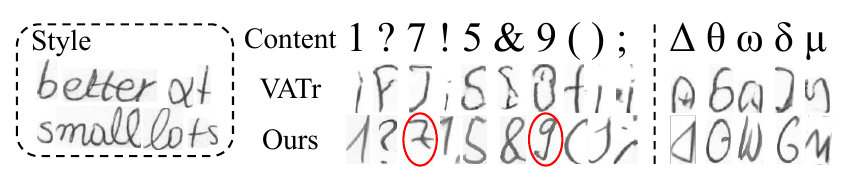}
    \caption{Failure cases. Generated numbers, punctuations, and Greek characters.}
    \label{fig:out-of-set}
    \vspace{-0.1in}
\end{figure}

We provide the generated long-tail characters (\ie digits and punctuations) and out-of-charset Greek letters from the IAM dataset by our One-DM in~\cref{fig:out-of-set}. One-DM occasionally struggles to generate long-tail characters with correct contents. For instance, as highlighted by red circles in~\cref{fig:out-of-set}, due to the relative scarcity of numbers $7$ and $9$ in the dataset, the model erroneously generates similar English letters $z$ and $g$, which are more common in the dataset. An effective solution is using an oversampling strategy to process long-tail characters during the training phase.

\begin{table}

 \caption{Effect of different filters on the IAM dataset.}
    \centering
    \includegraphics[width=0.9
    \linewidth]{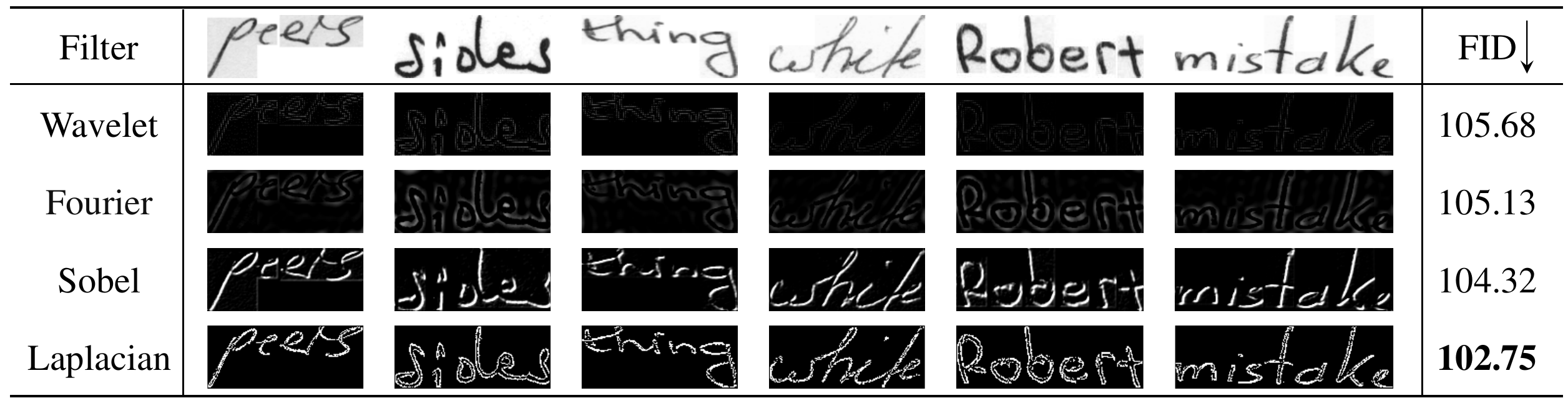}
    \label{filters}
  
 \end{table}

\section{More Ablation Experiments on IAM Dataset}\label{ablation-supp}
\begin{figure*}
  \centering
\includegraphics[width=0.9\linewidth]{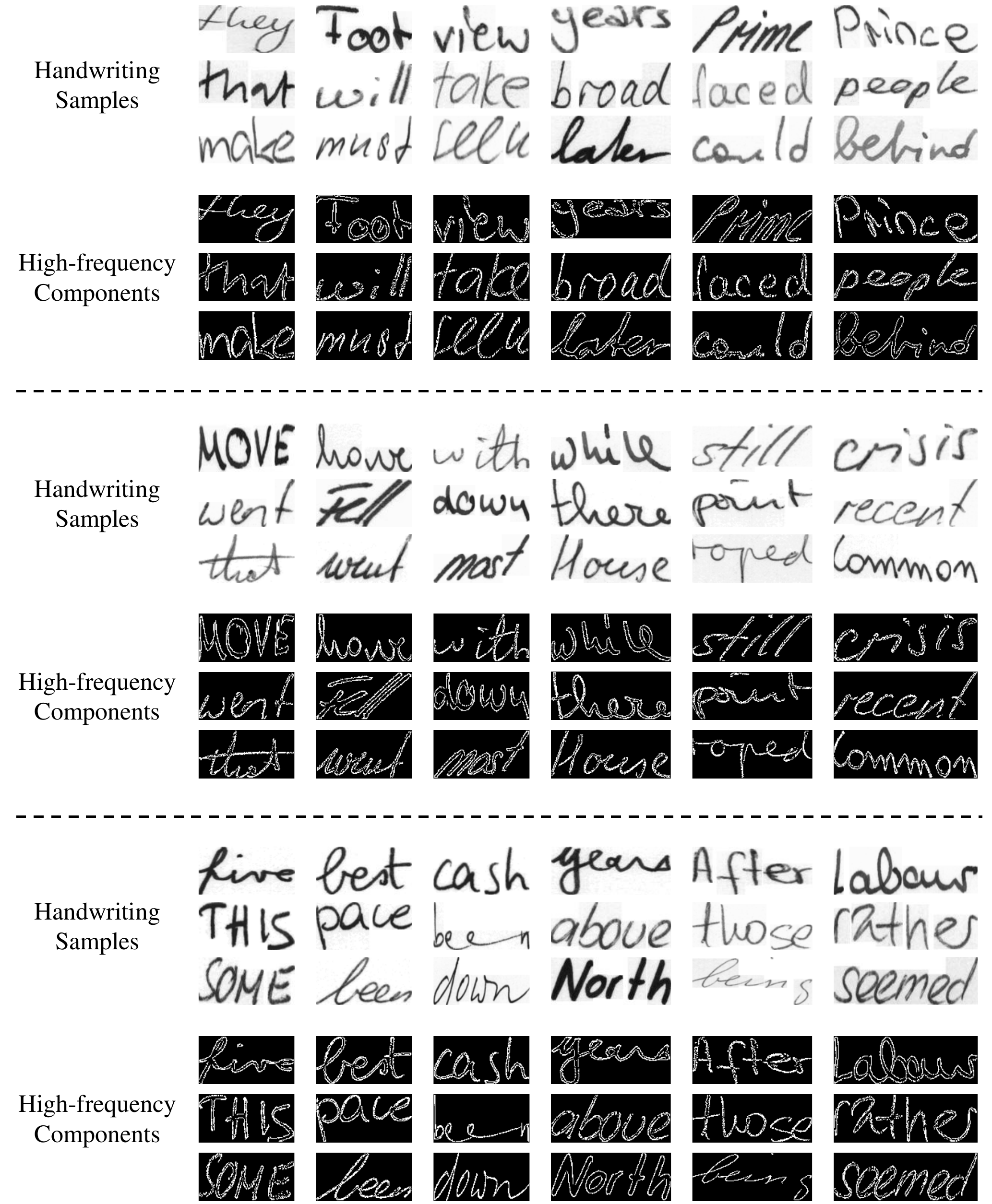}

   \caption{High-frequency information extracted by Laplacian filter.}
   \label{fig:vis_filter}

\end{figure*}

\textbf{Effect of different high-frequency filters.} We conduct ablation experiments to evaluate the effect of different high-frequency filters. As shown in~\cref{filters}, we find that Laplacian filter, compared to other filters, extracts more accurate and more complete character contours, achieving the best generation performance in terms of FID in OOV-U setting. More high-frequency results extracted by Laplacian filter are provided in~\cref{fig:vis_filter}. We can observe that our Laplacian filter can consistently extract high-frequency components in all samples.

\textbf{Discussion on different style-content fusion mechanisms.} The reason that we design the style-content fusion module based on cross-attention is because of its effectiveness in merging content and style to guide the diffusion model. Previous diffusion-based HTG methods~\cite{Ding2023ImprovingHO,icdar_NikolaidouRCSSSML23,zhu2023conditional} fuse content and style either by concatenation or by injecting style into the timestep embedding. However, they perform much worse than our style-content fusion module when ablating the fusion module (FID$\downarrow$: Ours-102.75 $vs.$ Ours with Concat-104.81 $vs.$ Ours with Timestep-104.26) under OOV-U setting in the IAM dataset.

\begin{figure}
    \centering
\includegraphics[width=0.65\linewidth]{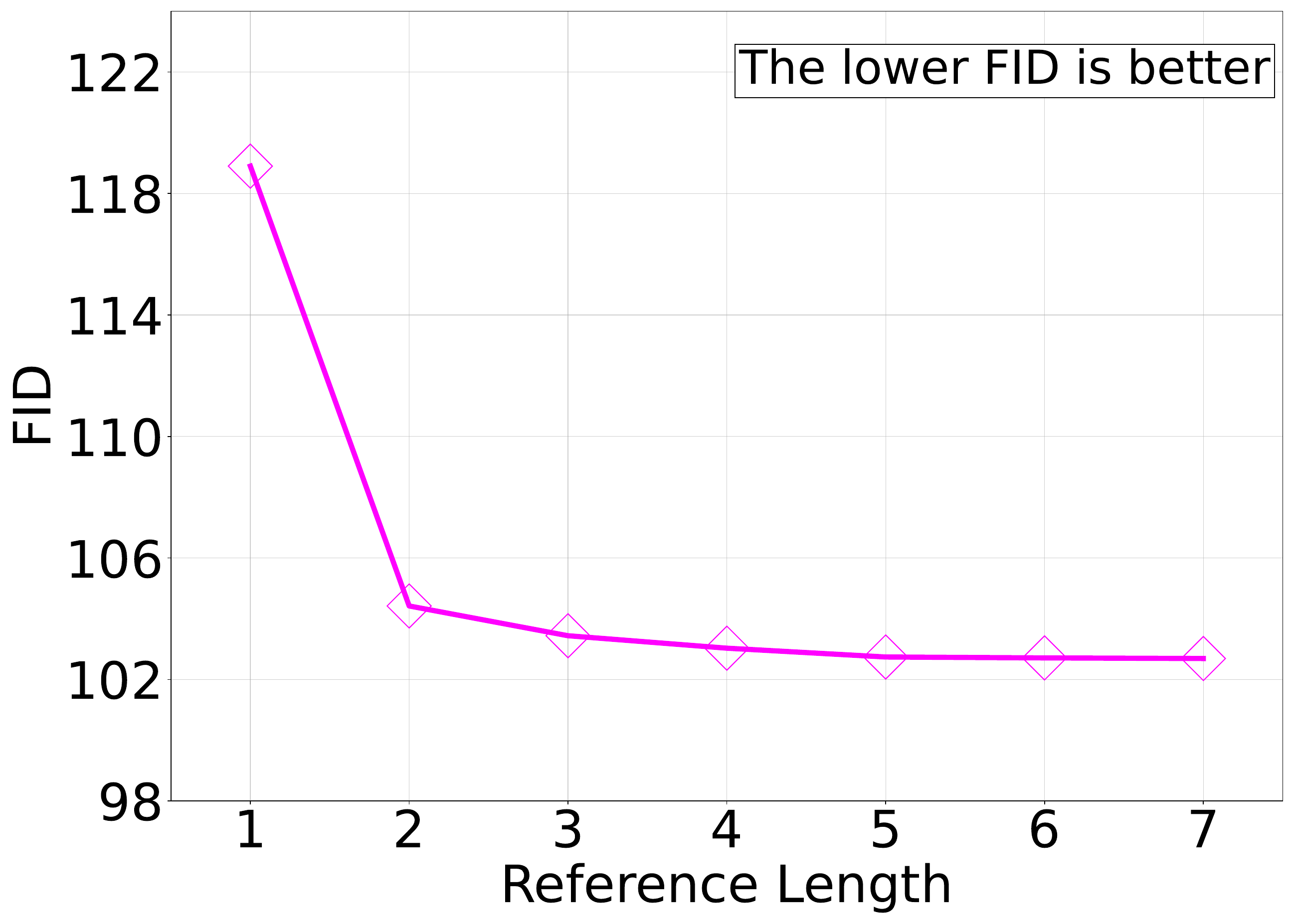}
    \caption{Effect of the length of style sample (i.e., the number of letters in a word) on the IAM dataset under the OOV-U setting. Here, we set the maximum length to 7, as 7 is the longest length among the handwritten samples provided by some writers in the IAM test set.}
    \label{ref_len}
    \vspace{-0.1in}
\end{figure}  

\textbf{Ablation on the Length of Style Reference.} We perform ablation experiments on the IAM dataset to explore how the length of the one-shot style reference (i.e., the number of letters in a word) affects the performance of our method. Specifically, during the generation process, we randomly select handwritten samples of specific lengths as style references and report the quantitative results under the OOV-U setting in~\cref{ref_len}. From these results, we observe that performance gradually improves as length increases. Importantly, when samples longer than four are used as style references, the model's performance tends to stabilize. Therefore, we recommend users provide style samples longer than four letters when using our generation model. It's important to emphasize that the model's performance is notably poor when the reference length is one. 
The reference sample of length one means a single letter and cannot provide style information like cursive connections, resulting in decreased performance. This can be alleviated by using samples longer than one as style references.

\section{Discussion on Different Ink Color}\label{ink}
\begin{figure}[t]
  \centering
\includegraphics[width=0.95\linewidth]{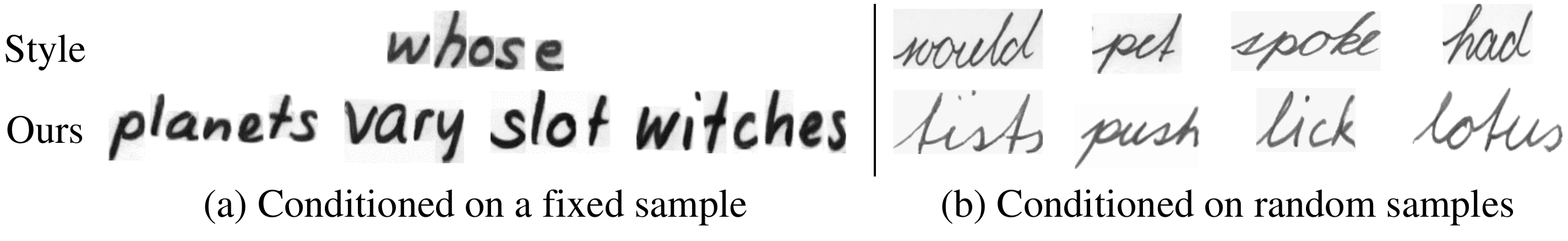}

   \caption{Generated results by our One-DM on IAM dataset.}
   \label{fig:sta}

\end{figure}

\begin{figure}[t]
  \centering
\includegraphics[width=0.75\linewidth]{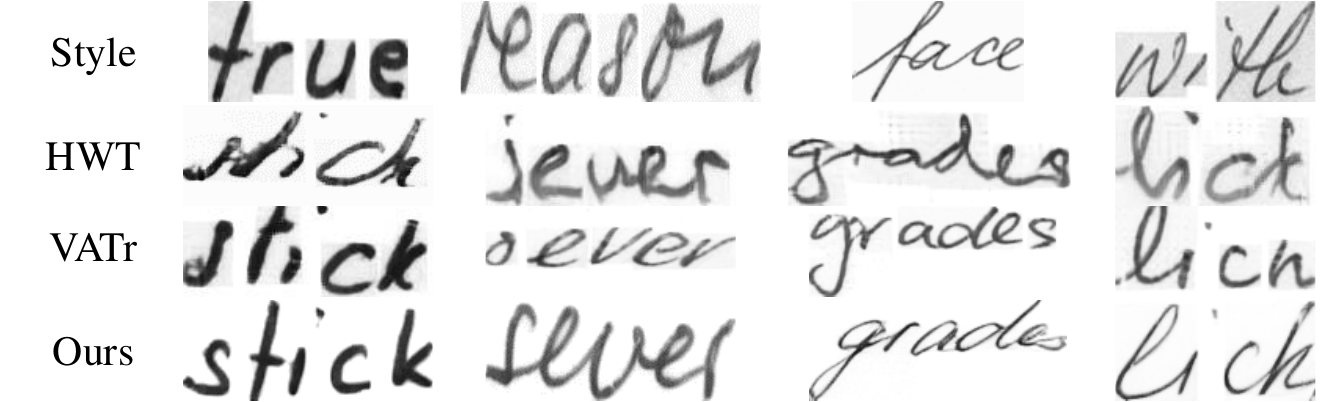}

   \caption{Comparisons with VATr and HWT on one-shot generation.}
   \label{fig:one-shot}

\end{figure}

As described in Section~\textcolor{red}{4.1}, our One-DM randomly selects a style reference from the target writer when generating each handwritten text. This leads our model to generate results with inconsistent ink colors since the sampled reference may have different ink colors (cf. the second and fourth columns of ``style examples" in  Figure \textcolor{red}{4} of Section~\textcolor{red}{4.2}). When conditioned on a fixed style reference or provided with consistent ink color samples, our model can generate uniform ink colors,  as shown in \cref{fig:sta}(a-b). In contrast, VATr and HWT generate color-consistent outputs in Figure \textcolor{red}{4} of Section~\textcolor{red}{4.2} since they require 15 reference samples to obtain an average style. However, when given only one reference sample, they tend to produce poor results, as shown in~\cref{fig:one-shot}.

\section{More Qualitative Comparisons}\label{ADD_SCR}
\subsection{English Handwritten Text Generation}\label{ADD_Eng}
\begin{figure*}
    \centering
    \includegraphics[width=0.9\linewidth]{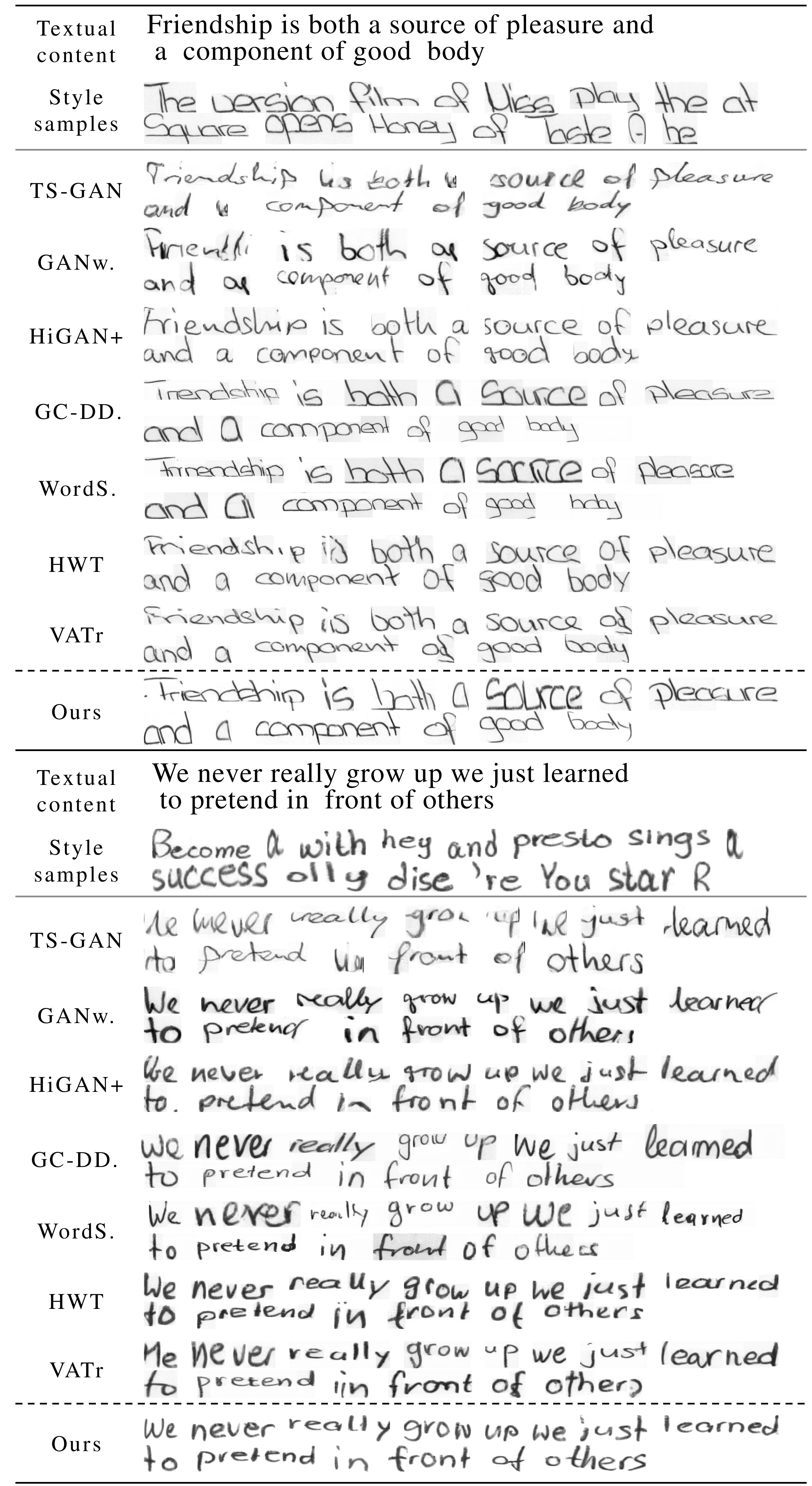}
    
    \caption{Additional comparisons between our method with state-of-the-art methods on English handwritten text generation.}
    \label{fig:English_1}
\end{figure*}

\begin{figure*}
    \centering
    \includegraphics[width=0.9\linewidth]{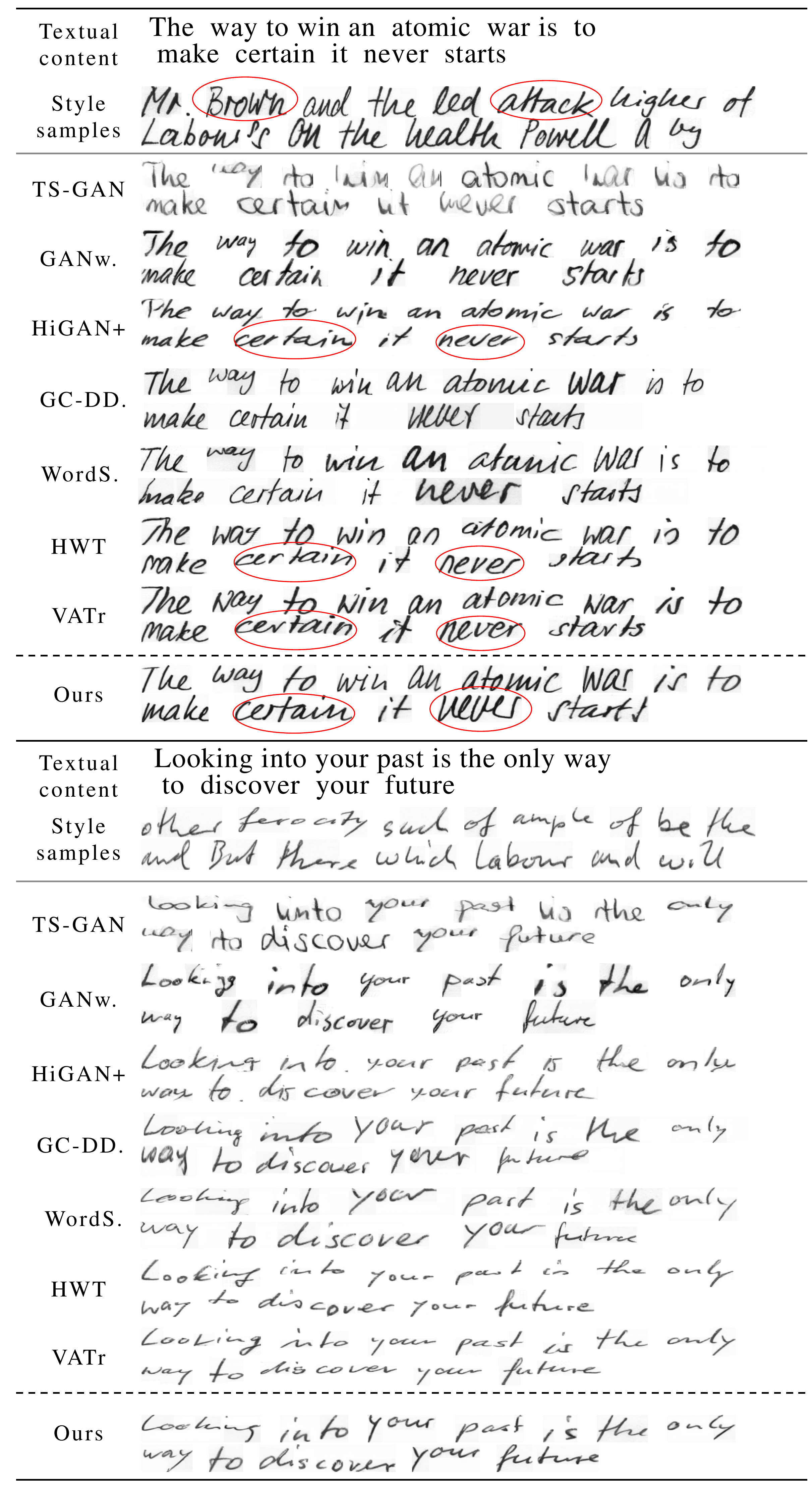}
    
    \caption{Additional comparisons between our method with state-of-the-art methods on English handwritten text generation.}
    \label{English_2}
\end{figure*}

\begin{figure*}
    \centering
    \includegraphics[width=0.9\linewidth]{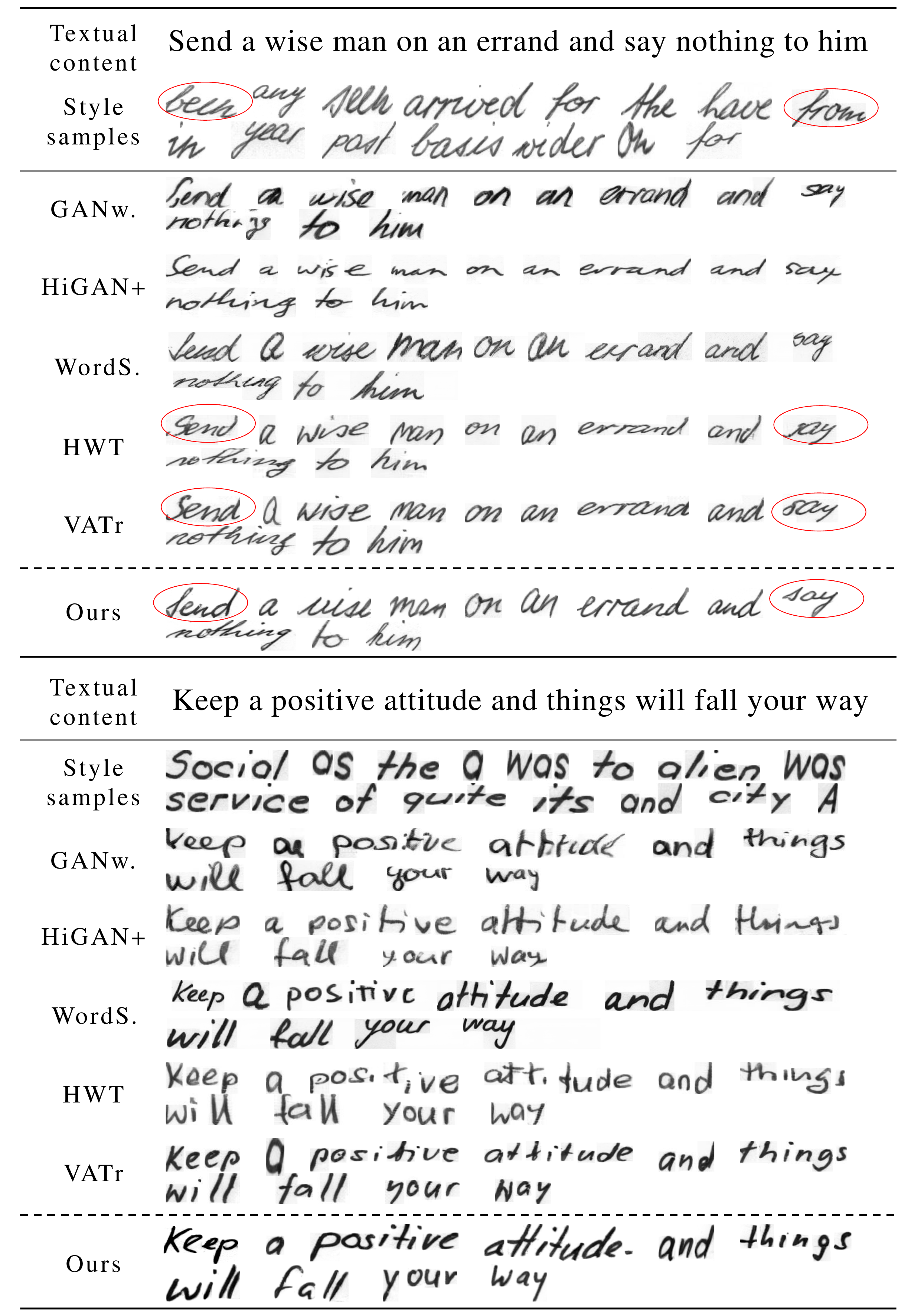}
    
    \caption{Additional comparisons between our method with state-of-the-art methods on English handwritten text generation.}
    \label{English_3}
\end{figure*}

\begin{figure*}
    \centering
    \includegraphics[width=0.9\linewidth]{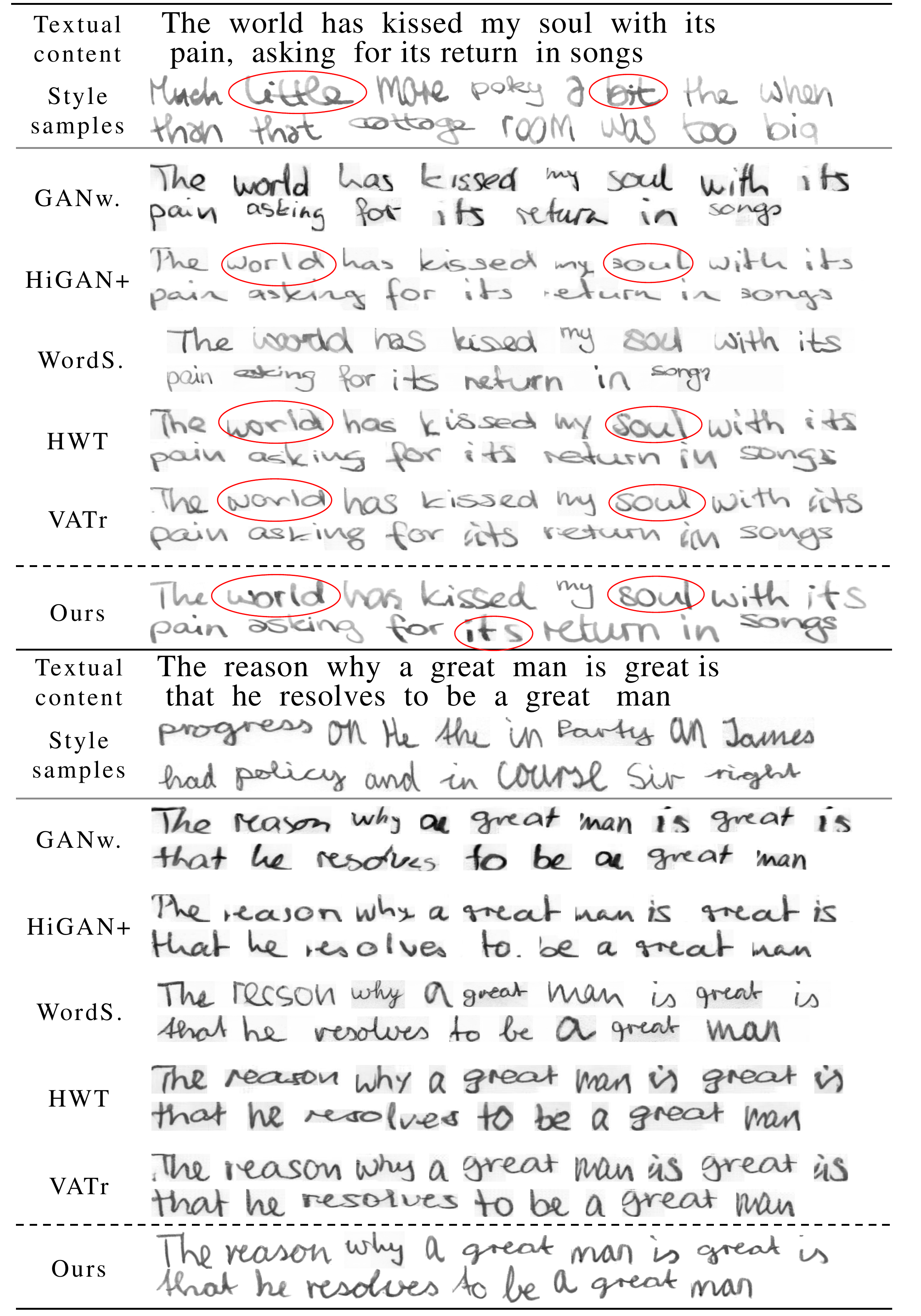}
    \caption{Additional comparisons between our method with state-of-the-art methods on English handwritten text generation.}
    \label{English_4}
\end{figure*}

From~\cref{fig:English_1} to~\cref{English_4}, we show more qualitative comparisons between our proposed One-DM and the previous state-of-the-art works, i.e., TS-GAN~\cite{DavisMPTWJ20}, GANwriting~\cite{kang2020ganwriting}, HiGAN$+$\cite{gan2022higan+}, GC-DDPM~\cite{Ding2023ImprovingHO}, WordStylist~\cite{icdar_NikolaidouRCSSSML23}, HWT~\cite{bhunia2021handwriting} and VATr~\cite{pippi2023handwritten}. To ensure a fair comparison,  we utilize identical textual content for all methods, as displayed in the rows ``Textual content'' of each figure. Each figure's rows of ``Style samples'' showcase style examples from different writers. Subsequent rows are dedicated to showcasing the generated results from competitors and our One-DM. \cref{fig:English_1} shows that TS-GAN and GANwriting struggle to capture the character slant of the reference samples, while diffusion-based methods (i.e., GC-DDPM and WordStylist) tend to produce results with noticeable background noise. The cursive joins in generated samples by HiGAN$+$, HWT, and VATr occasionally lack realism, as shown in~\cref{English_2}. VATr and HWT sometimes fail to accurately capture the initial stroke details of characters, as evident from~\cref{English_3}. More interestingly, from ~\cref{English_4}, we can observe that the internal colors of strokes generated by HiGAN$+$, HWT, and VATr are typically uniform. Conversely, our method is capable of more realistically emulating the gradient effects of stroke colors found in real samples.

\subsection{Chinese Handwriting Generation}\label{Chinese_qua}
\begin{figure*}
    \centering
    \includegraphics[width=0.9\linewidth]{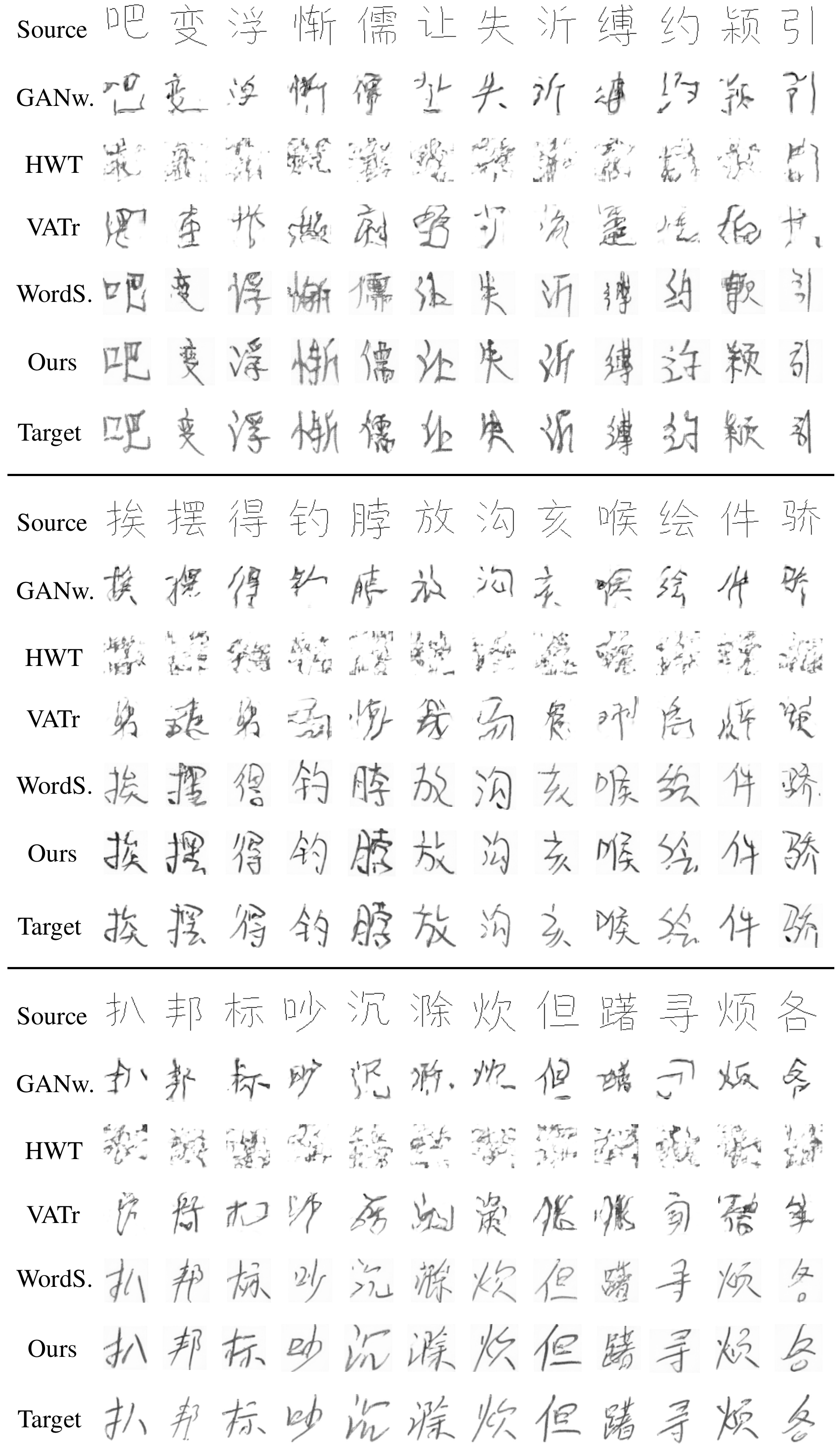}
    
    \caption{Additional comparisons between our method with state-of-the-art methods on Chinese handwriting generation.}
    \label{Chinese_character_1}
\end{figure*}

\begin{figure*}
    \centering
    \includegraphics[width=0.9\linewidth]{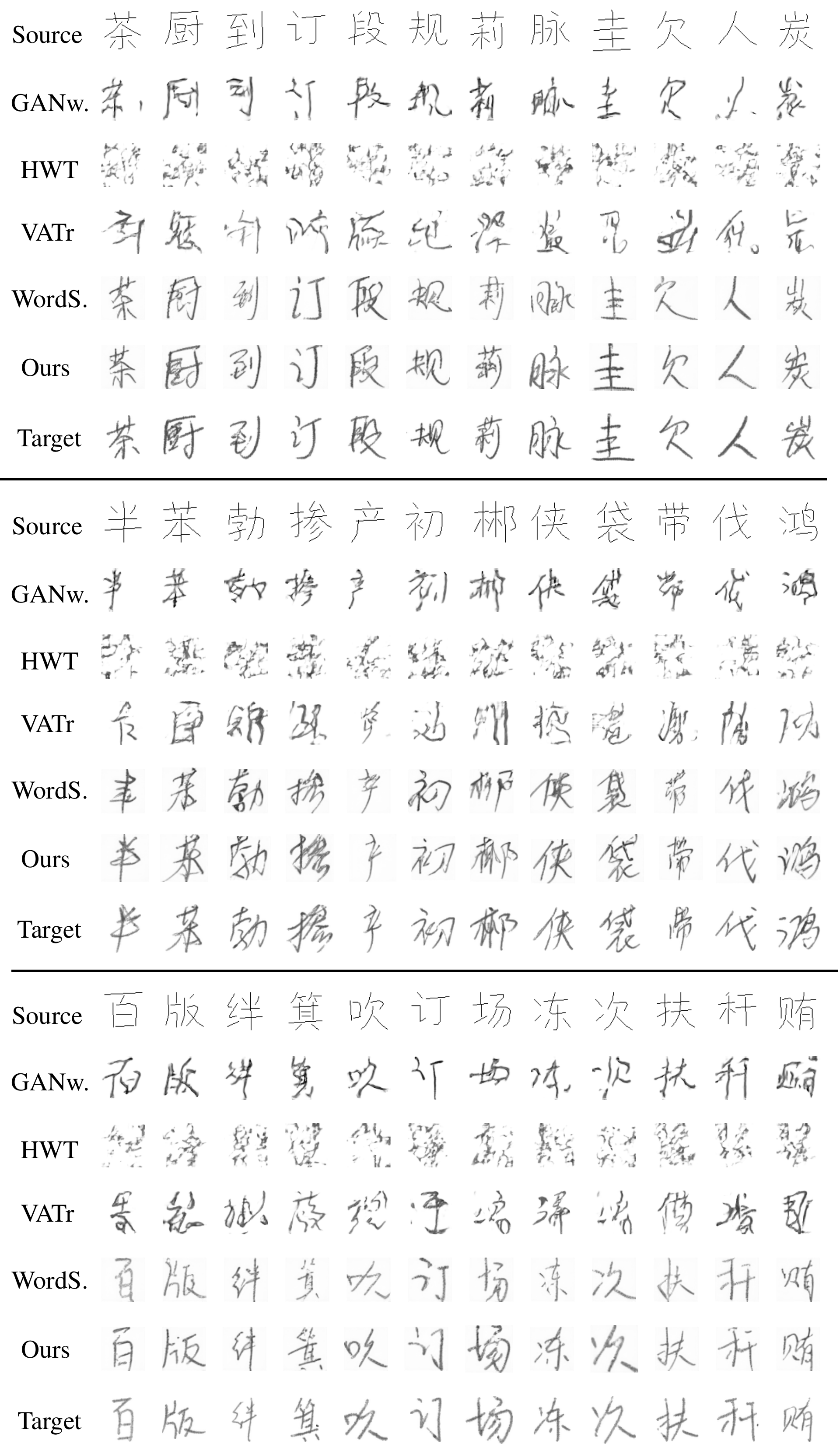}
   
    \caption{Additional comparisons between our method with state-of-the-art methods on Chinese handwriting generation.}
     \label{Chinese_character_2}
\end{figure*}

\begin{figure*}
    \centering
    \includegraphics[width=0.9\linewidth]{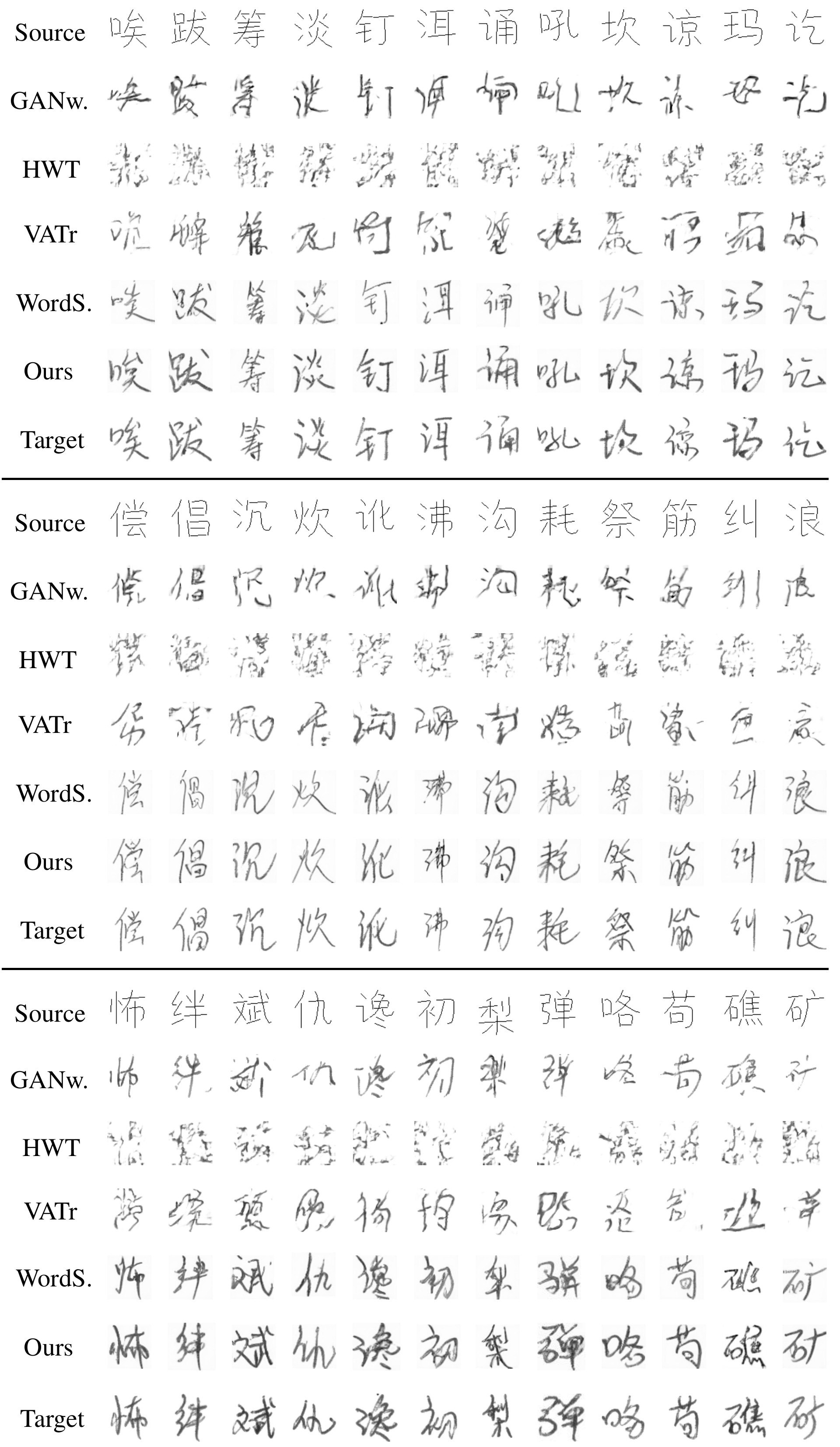}
    
    \caption{Additional comparisons between our method with state-of-the-art methods on Chinese handwriting generation.}
    \label{Chinese_character_3}
\end{figure*}

We provide more qualitative comparisons on handwritten Chinese character generation with GANwriting~\cite{kang2020ganwriting}, HWT~\cite{bhunia2021handwriting}, VATr~\cite{pippi2023handwritten}, and WordStylist~\cite{icdar_NikolaidouRCSSSML23} in~\cref{Chinese_character_1} to~\cref{Chinese_character_3}. 
The ``Source'' rows in each figure display standard Chinese characters varying in content,  while each ``Target'' row showcases real characters from the same writer. These results clearly show that the handwritten samples synthesized by our method (rows of ``Ours'')  are the most competitive in terms of style imitation (e.g., stroke color) and character structure preservation.

\subsection{Japanese Handwriting Generation}\label{Japan_qua}
\begin{figure*}
    \centering
    \includegraphics[width=0.9\linewidth]{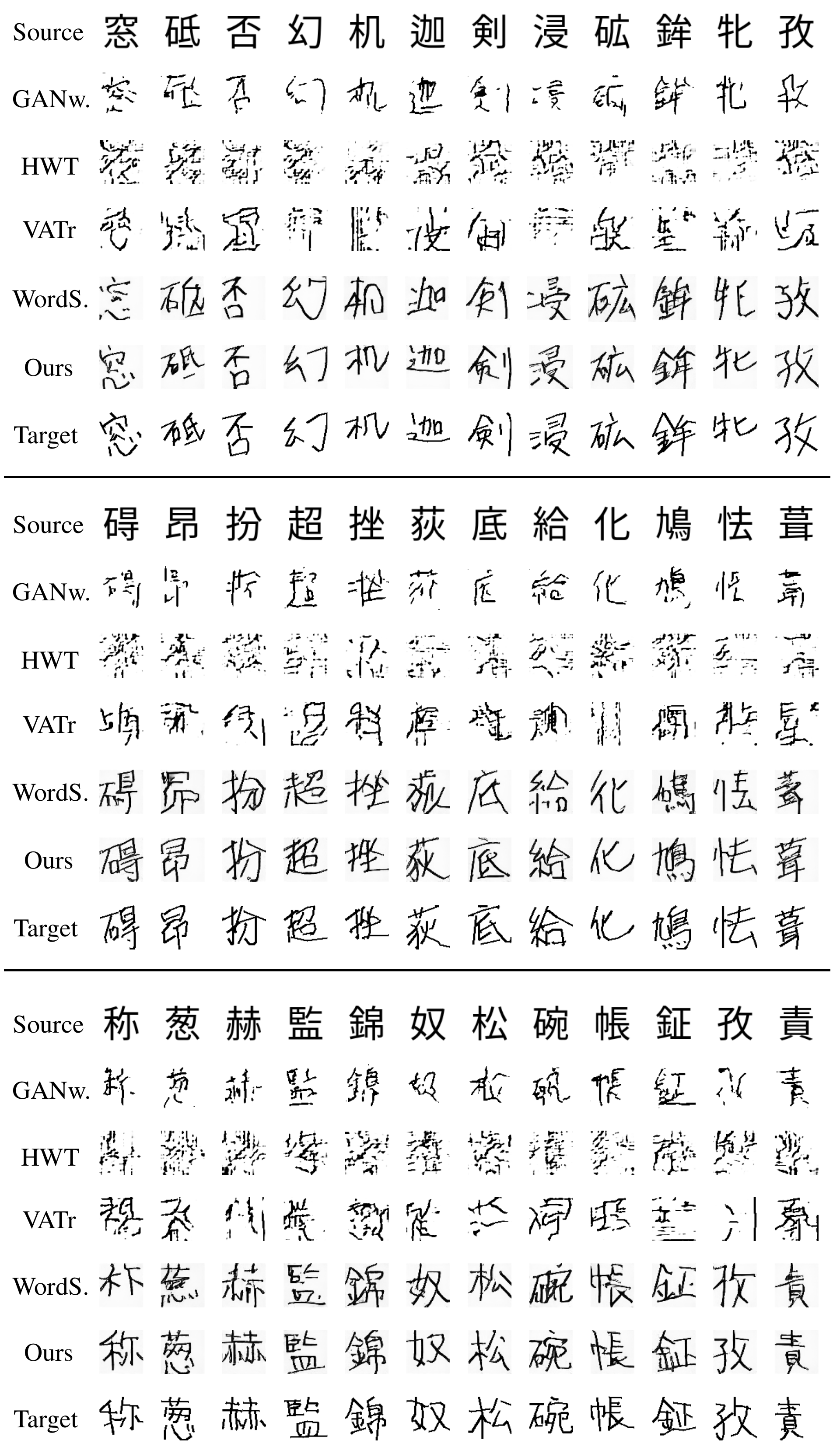}
    
    \caption{Additional comparisons between our method with GANwriting, HWT, VATr, and WordStylist on Japanese handwriting generation.}
    \label{Japanese_character_1}
\end{figure*}

\begin{figure*}
    \centering
    \includegraphics[width=0.9\linewidth]{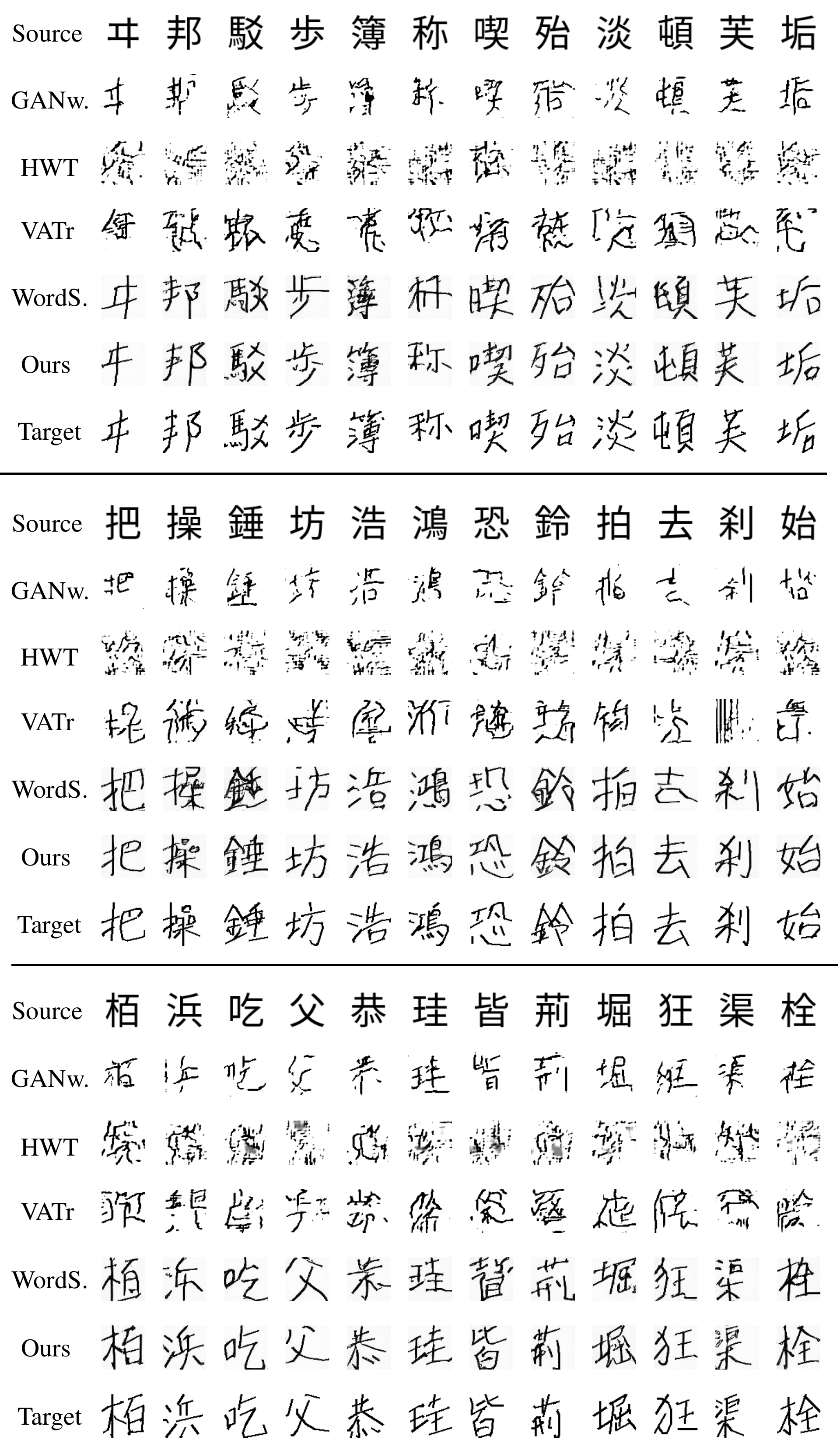}
    
    \caption{Additional comparisons between our method with GANwriting, HWT, VATr, and WordStylist on Japanese handwriting generation.}
    \label{Japanese_character_2}
\end{figure*}

\begin{figure*}
    \centering
    \includegraphics[width=0.9\linewidth]{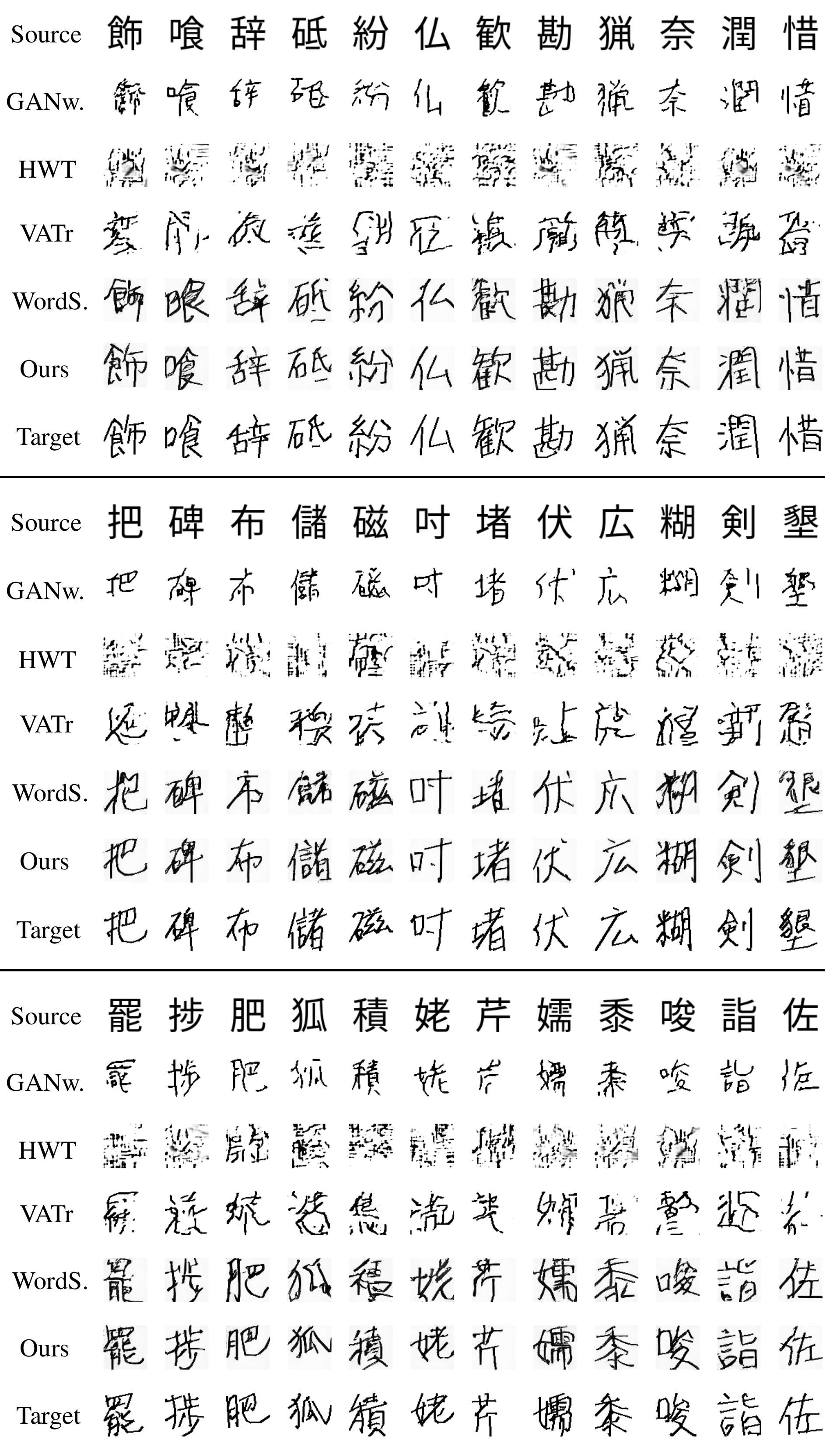}
    
    \caption{Additional comparisons between our method with GANwriting, HWT, VATr, and WordStylist on Japanese handwriting generation.}
    \label{Japanese_character_3}
\end{figure*}

In~\cref{Japanese_character_1} to~\cref{Japanese_character_3}, we show more visual results with a comparison to GANwriting~\cite{kang2020ganwriting}, HWT~\cite{bhunia2021handwriting}, VATr~\cite{pippi2023handwritten}, and WordStylist~\cite{icdar_NikolaidouRCSSSML23} on Japanese handwriting generation. Similarly, in each figure, the ``Source'' rows feature Japanese standard characters with diverse content, whereas each ``Target'' row exhibits real characters from the same author. From these results, we can observe that GANwriting (rows of ``GANw.'') often produces characters that are incomplete and prone to missing strokes. HWT and VATr, on the other hand, generate characters that are very blurry, with the content of the characters being distorted. WordStylist struggles to generate stroke details consistent with the target sample. Compared with the competitors, our method (rows of ``Ours'') accurately reproduces the complete structure of handwritten Japanese characters (particularly the stroke details) and replicates the style patterns of real characters. 

\end{document}